\documentclass[runningheads]{llncs}

\usepackage[year=2026]{eccv}
\usepackage{eccvabbrv}

\usepackage{graphicx}
\usepackage{booktabs}

\usepackage[accsupp]{axessibility}  % Improves PDF readability for those with disabilities.

\usepackage{hyperref}

\usepackage{orcidlink}

\usepackage{marvosym}
\usepackage{booktabs}
\usepackage{graphicx}
\usepackage{algorithm}
\usepackage{algpseudocode} 
\usepackage{kotex}
\usepackage{multirow}
\usepackage{orcidlink}
\usepackage{placeins}
\usepackage{afterpage}
\usepackage{wrapfig}
\usepackage{amsfonts}
\usepackage{nicefrac} 
\usepackage{microtype}
\usepackage{xcolor} 
\usepackage{colortbl}
\usepackage{color}
\usepackage{amsmath}
\usepackage{caption}
\usepackage{float}
\definecolor{first}{HTML}{547CB1} %
\definecolor{improve}{HTML}{1E73C4} %

\usepackage{enumitem}
\usepackage{wrapfig}

\begin{document}

\title{LivingWorld: Interactive 4D World Generation with Environmental Dynamics}

\titlerunning{LivingWorld}

\vspace{-2mm}
\author{
Hyeongju Mun\textsuperscript{*}\orcidlink{0009-0008-5289-5535} \and
In-Hwan Jin\textsuperscript{*}\orcidlink{0009-0008-9202-6510} \and
Sohyeong Kim\orcidlink{0009-0001-5755-9820} \and
Kyeongbo Kong\textsuperscript{\Letter}\orcidlink{0000-0002-1135-7502}
}

\authorrunning{H. Mun, I. Jin et al.}

\vspace{-3mm}
\institute{
Department of Electrical and Electronic Engineering, Pusan National University\\[2pt]
\email{\{201924128, ihjin, soh9119, kbkong\}@pusan.ac.kr}\\[4pt]
Project page: \url{https://paper.pnu-cvsp.com/LivingWorld/}
}

\maketitle

\begingroup
\renewcommand{\thefootnote}{}
\footnotetext{\textsuperscript{*} Equal contribution. \quad \textsuperscript{\Letter} Corresponding author.}
\endgroup

\vspace{-8.5mm}

\begin{figure}[H]
\centering
\includegraphics[width=\textwidth]{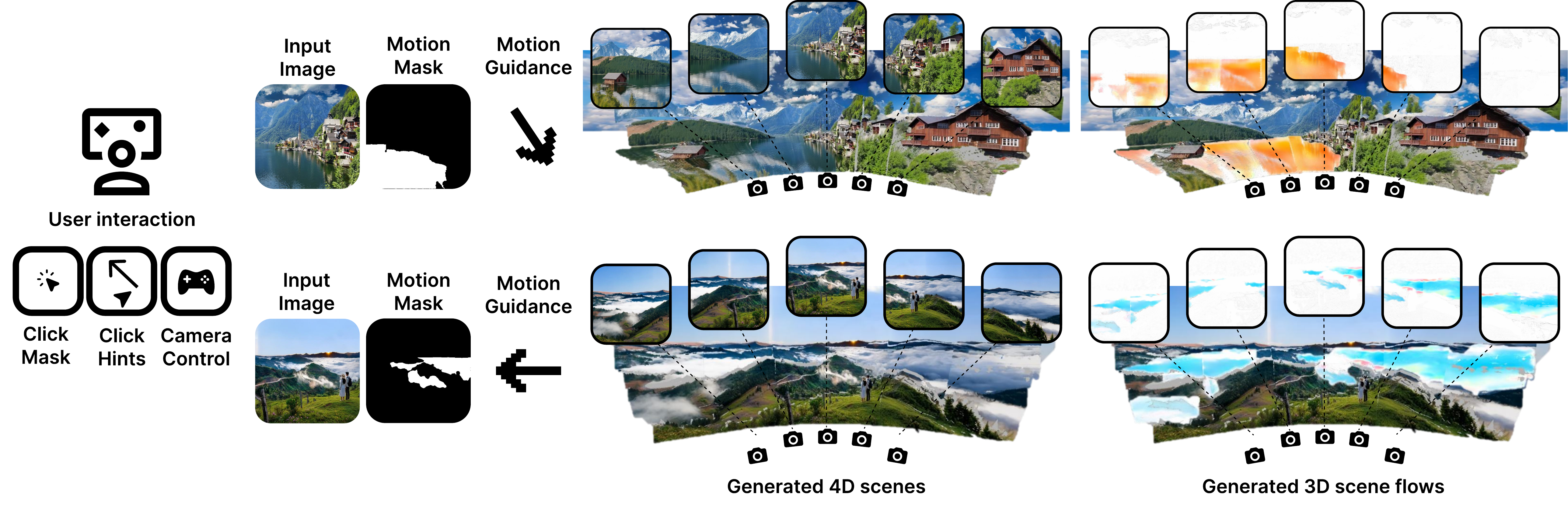}

\begin{minipage}{0.9\textwidth}
\caption{LivingWorld generates a dynamic 4D world with environmental dynamics from a single image. Our geometry-aware alignment module maintains globally coherent scene dynamics as the world progressively expands.}
\label{fig:teaser}
\end{minipage}
\end{figure}

\vspace{-16mm}
\begin{abstract}
We introduce \textbf{LivingWorld}, an interactive framework for generating 4D worlds with environmental dynamics from a single image. While recent advances in 3D scene generation enable large-scale environment creation, most approaches focus primarily on reconstructing static geometry, leaving scene-scale environmental dynamics such as clouds, water, or smoke largely unexplored.
Modeling such dynamics is challenging because motion must remain coherent across an expanding scene while supporting low-latency user feedback.
LivingWorld addresses this challenge by progressively constructing a globally coherent motion field as the scene expands.
To maintain global consistency during expansion, we introduce a geometry-aware alignment module that resolves directional and scale ambiguities across views. We further represent motion using a compact hash-based motion field, enabling efficient querying and stable propagation of dynamics throughout the scene.
This representation also supports bidirectional motion propagation during rendering, producing long and temporally coherent 4D sequences without relying on expensive video-based refinement. On a single RTX 5090 GPU, generating each new scene expansion step requires 9 seconds, followed by 3 seconds for motion alignment and motion field updates, enabling interactive 4D world generation with globally coherent environmental dynamics. Video demonstrations are available at \href{https://paper.pnu-cvsp.com/LivingWorld/}
{\textcolor{magenta}{\texttt{paper.pnu-cvsp.com/LivingWorld}}}.

\vspace{-2mm}
\end{abstract}

\section{Introduction}
\label{sec:intro}
Recent advances in 3D representation~\cite{kerbl20233d, mildenhall2021nerf} and generative modeling~\cite{baldridge2024imagen, wan2025wan} have enabled the interactive creation of large-scale virtual environments from minimal visual input. Such systems~\cite{yu2025wonderworld} allow users to construct and manipulate complex scenes in real time, supporting open-ended world generation and significantly lowering the barrier to 3D content creation. However, most existing approaches generate worlds that remain fundamentally static, focusing primarily on reconstructing geometry and appearance while leaving the environment itself motionless.  

In many real-world environments, certain environmental processes are inherently dynamic. Rivers flow, waves propagate along coastlines, and clouds drift across the sky. These phenomena are not merely moving objects within a scene but environmental dynamics that are intrinsically tied to the scene itself. Ignoring such dynamics leads to environments that may appear visually plausible yet fail to faithfully reflect how real-world environments behave. This limitation becomes increasingly significant as interactive world generation is explored in applications such as perception, simulation, and embodied intelligence, where environments must capture not only appearance but also dynamic characteristics.

However, modeling environmental dynamics in interactive world generation remains challenging. Such motions often span large spatial regions and evolve continuously over time. As scenes expand progressively, motion guidance must be determined while accounting for how previously generated dynamics influence the evolving environment. 
This creates a feedback-driven process in which motion must remain globally coherent while being updated interactively. To support such interaction, the motion representation must maintain spatial and temporal consistency across the scene while allowing efficient updates so that users can observe the resulting dynamics and refine their guidance in real time.

A closely related line of research explores controllable video generation~\cite{wiedemer2025video, zhang2025tora, yang2024cogvideox}, where camera trajectories or motion controls are used to synthesize temporally coherent image sequences. 
While these approaches can produce visually compelling videos, they typically generate image sequences without constructing an explicit and persistent 3D world representation. 
As a result, maintaining geometrically consistent dynamics across viewpoints becomes difficult, particularly when the scene is expanded or revisited from new viewpoints.
Another line of research~\cite{jin2025optimizing} models motion using Eulerian representations~\cite{holynski2021animating, mahapatra2023text, choi2024stylecinegan}, originally developed for optical flow. 
These approaches extend Eulerian motion models to 4D scene generation by associating motion with spatial points in 3D space. 
Although they enforce geometric consistency through multi-view flow estimation, they often rely on iterative refinement driven by video supervision, where motion is optimized indirectly through appearance reconstruction. 
Such optimization-based pipelines introduce significant computational overhead and reduce responsiveness in interactive settings, making it difficult to provide immediate visual feedback during scene creation.

To address these limitations, we propose \textbf{LivingWorld}, an interactive framework that constructs environmental dynamics within an explicit 3D scene representation (Fig.~\ref{fig:teaser}). 
Rather than recovering motion through video-driven appearance reconstruction, LivingWorld represents environmental dynamics using a global motion field, a continuous motion representation defined over the entire reconstructed scene, and progressively updates it as the scene expands.
Given user-provided motion prompts, the system estimates motion cues from newly generated views and consolidates them into a unified global motion field.
To maintain consistency across views, we introduce a geometry-aware alignment module that helps resolve directional and scale ambiguities in 3D space. 
We further represent motion using a compact hash-based motion field, enabling efficient querying and stable propagation of motion throughout the scene. 
This representation also supports bidirectional motion propagation during rendering, allowing long and temporally coherent 4D sequences to be produced without additional video-driven refinement. 
Together, these design choices enable globally consistent environmental dynamics to be constructed within seconds, enabling interactive 4D world generation with fast visual feedback. 
While our approach focuses on environmental motion that governs large-scale scene dynamics, it establishes a scene-level dynamic foundation onto which foreground object motions can be integrated using object-centric motion models.
Our contributions are summarized as follows:

\begin{itemize}
\item We introduce \textbf{LivingWorld}, an interactive framework for generating 4D worlds from a single image that models environmental dynamics as a global motion field.

\item We propose a \textbf{geometry-aware alignment module} that consolidates motion cues from newly generated views into a globally consistent 3D motion representation.

\item We design a \textbf{hash-based motion field with bidirectional motion propagation}, enabling fast motion field construction and stable spatio-temporal motion propagation across the scene.

\end{itemize}

\vspace{-2mm}
\section{Related Works}
\label{sec:relatedworks}
\vspace{-2mm}
\subsection{3D Scene Generation}
\vspace{-2mm}
The goal of 3D scene generation is to synthesize explorable and coherent environments from limited inputs such as single images or text prompts.
Early methods focused on expanding static scenes from a single image~\cite{kaneva2010infinite, liu2021infinite}. With the emergence of diffusion-based generative models, recent work has progressed toward world-scale scene generation and procedural environment construction. Some approaches aim to directly synthesize expansive environments through terrain generation and structured layout modeling~\cite{wu2024blockfusion, hua2025sat2city, zhou2025scenex}.

Another dominant paradigm follows a render–refine–repeat pipeline, where scenes are iteratively rendered, refined using depth alignment, and updated with newly predicted geometry. Text2Room~\cite{hollein2023text2room} applies this strategy to generate mesh-based indoor environments from text prompts. LucidDreamer~\cite{chung2023luciddreamer} and Text2Immersion~\cite{ouyang2023text2immersion} extend this paradigm to realistic indoor-to-outdoor environments using Gaussian Splatting~\cite{kerbl20233d}. WonderJourney~\cite{yu2024wonderjourney} further incorporates large language models to generate semantically diverse and interconnected scenes. WonderWorld~\cite{yu2025wonderworld} introduces an interactive framework that allows users to progressively construct large-scale 3D environments.
Despite these advances, existing approaches primarily focus on reconstructing static geometry.  While such representations can produce visually plausible environments, they do not explicitly model environmental motion, causing the generated worlds to remain static. As a result, incorporating temporally coherent environmental dynamics into interactive 3D environments remains an open challenge, motivating the need for interactive 4D world generation.

\vspace{-3mm}
\subsection{Single Image Animation}
\vspace{-2mm}
Single-image animation aims to generate dynamic visual effects from static images, particularly for natural phenomena such as clouds, water, or smoke. Early approaches relied on manual layer decomposition or simple physical models~\cite{chuang2005animating, jhou2015animating}. Later works adopted deep neural networks to infer motion fields directly from static images~\cite{endo2019animating, logacheva2020deeplandscape}. Holynski et al.~\cite{holynski2021animating} proposed an Eulerian-based model that predicts dense looping flow fields for realistic image animations. Subsequent research introduced controllable motion generation through semantic guidance. Text2Cinemagraph~\cite{mahapatra2023text} and StyleCineGAN~\cite{choi2024stylecinegan} enable text-driven or style-guided motion synthesis, while diffusion-based approaches further improve temporal realism and controllability~\cite{shi2024motion, xing2025motioncanvas, shi2025motionstone, jin2025optimizing}.
Recent diffusion-based motion generation models~\cite{wiedemer2025video, zhang2025tora, yang2024cogvideox} can synthesize motion from static images, producing visually rich animations with plausible environmental dynamics and camera movement.
However, these approaches synthesize image sequences without maintaining an explicit and persistent 3D scene representation, making it difficult to ensure geometric consistency across viewpoints or during scene expansion.
To address this limitation, several works attempt to incorporate 3D structure for improved spatial consistency.
3D-Cinemagraphy~\cite{li20233d} and Make-It-4D~\cite{shen2023make} infer pseudo 3D scene flow from depth-based layer decomposition, which often leads to geometric distortions under large viewpoint changes.
4DGS-Cinemagraphy~\cite{jin2025optimizing} introduces a Gaussian Splatting representation to maintain multi-view consistency, but relies on rendering-based optimization to recover motion, resulting in substantial computational overhead.

\vspace{-3mm}
\subsection{4D Scene Generation}
\vspace{-2mm}
Generating temporally coherent and geometrically consistent 4D scenes remains a challenging problem in computer vision and graphics. Early works extended static 3D representations into dynamic sequences by transferring motion priors from video diffusion models. MAV3D~\cite{singer2023text} introduced a hybrid score distillation framework to inject temporal motion into 3D assets, while subsequent studies improved motion fidelity through trajectory-aware optimization and refined score-distillation strategies~\cite{bahmani20244d, zheng2024unified, bahmani2024tc4d, zeng2024trans4d}. Although these approaches produce visually compelling results, they typically focus on object-centric or bounded scenes and struggle to maintain stable dynamics under large viewpoint changes or long camera trajectories.

More recent works attempt to extend diffusion-based generation to dynamic 4D scenes with explicit scene representations. CAT4D~\cite{wu2025cat4d} leverages a multi-view video diffusion model to synthesize temporally consistent multi-view videos from monocular video inputs, which are then used to optimize a 4D Gaussian Splatting representation. Another line of work explores hybrid frameworks that combine physical simulation with generative models. WonderPlay~\cite{li2025wonderplay} and PerpetualWonder~\cite{zhan2026perpetualwonder} reconstruct a coarse 3D scene from a single image and generate physical states using a simulator. These states are used as conditioning signals for a video generation model to synthesize multi-view videos, which subsequently supervise the optimization of dynamic Gaussian representations.
Despite their different formulations, these approaches ultimately rely on generated videos as supervision to infer motion in 3D. Such video-driven optimization introduces significant computational overhead and limits the ability to incorporate user feedback in interactive environments. In contrast, our method directly constructs environmental motion within a continuous 3D scene representation without relying on video supervision. This design enables rapid updates of global motion fields, allowing user feedback to be incorporated efficiently during interactive world generation.

\vspace{-4pt}

\begin{figure*}[t]
    \centering
    \includegraphics[width=\textwidth]{figure/overframe.pdf}
    \caption{\textbf{Overall framework of the proposed method.}
    Starting from a single input image, our framework first constructs an initial 4D scene by estimating user-guided motion cues and encoding them into a hash-based global motion field.
    As the scene progressively expands through camera movement and outpainting, newly estimated motion is aligned with previously accumulated motion using the proposed geometry-aware alignment module to maintain globally coherent environmental dynamics.
    During rendering, bidirectional motion propagation with an opacity scheduler produces temporally stable and seamless environmental dynamics.}
    \label{fig:framework}
        \vspace{-1.5mm}
\end{figure*}

\vspace{-2mm}
\section{Method}
\vspace{-3mm}

Sec.~\ref{overview} first provides an overview of the proposed framework.
Sec.~\ref{global_aling} then describes how the global motion field is progressively constructed from user-guided motion cues.
Finally, Sec.~\ref{motion_prop} presents motion propagation and rendering for generating temporally coherent 4D scenes.

\vspace{-3mm}
\subsection{Overview}
\vspace{-2mm}
\label{overview}
% \vspace{-3mm}
Starting from a single input image, an initial 4D scene is generated and progressively expanded as new views are synthesized. Camera poses $C_{\text{gen}}$ and text prompts $\mathcal{U}$ guide the spatial and semantic expansion of the scene, while sparse motion cues specify regions that should exhibit dynamic behavior.
As illustrated in Fig.~\ref{fig:framework}, the framework first constructs an initial 4D scene and then progressively expands it through user-guided camera movement and outpainting. Throughout both stages, a global motion field is initialized and continuously updated to maintain globally coherent environmental dynamics. During rendering, the learned motion field is queried for bidirectional motion propagation with an opacity scheduler, producing temporally coherent 4D scenes.

\vspace{-3mm}
\subsection{Motion Field Generation}
\vspace{-2mm}
\label{global_aling}
The global motion field is progressively constructed throughout the initial and interactive 4D scene generation stages. Specifically, the proposed motion field generation consists of three components: Eulerian motion estimation, motion alignment, and motion field learning.

\noindent\textbf{Eulerian Motion Estimation.}
Given an input image or a newly generated view, we estimate motion cues using an Eulerian flow representation~\cite{holynski2021animating}, which is particularly effective for modeling fluid-like environmental dynamics. Unlike Lagrangian approaches that explicitly track the trajectories of particles or primitives, Eulerian flow assigns a velocity vector to each spatial location.

Given an image $I$, the model predicts a per-pixel flow field
$M_t(\mathbf{u})$ that represents the instantaneous velocity
at image coordinate $\mathbf{u}\in\mathbb{R}^2$ and time $t$.
Future positions are obtained via Euler integration,
\begin{equation}
\label{eq:euler_motion}
\mathbf{u}_{t+1} = \mathbf{u}_t + M_t(\mathbf{u}_t),
\end{equation}
and the cumulative flow is recursively computed as
\begin{equation}
\label{eq:euler_integral}
F_{0 \rightarrow t}(\mathbf{u}_0) = F_{0 \rightarrow t-1}(\mathbf{u}_0) + M_t(\mathbf{u}_0 + F_{0 \rightarrow t-1}(\mathbf{u}_0)),
\end{equation}
where $F_{0 \rightarrow t}(\cdot)$ denotes the accumulated flow from time $0$ to $t$.
This formulation enables motion to be estimated directly in image space 
using only velocity predictions.
We denote the Eulerian flow predictor as $\mathrm{EF}(\cdot)$ and apply it 
to each generated view $I_i$ to obtain a 2D flow field
\begin{equation}
F_i = \mathrm{EF}(I_i).
\end{equation}

In interactive scenarios, users optionally specify motion regions through seed points
$\{p_i\}$.
These points are converted into binary masks $\{m_i\}$ using SAM~\cite{kirillov2023segment}, while directional hints $\{h_i\}$ indicate the desired motion orientation.
The resulting view-wise flow fields provide motion cues for the subsequent motion alignment stage.

\noindent\textbf{Motion Alignment.}
The estimated view-wise flow fields $F_i$ are lifted into 3D using the predicted depth maps.
For each pixel $\mathbf{u}$ with depth $D_i(\mathbf{u})$, we unproject its image coordinate into the world coordinate system and associate the corresponding 2D flow vector $F_i(\mathbf{u})$ with its recovered 3D position $\mathbf{x}\in\mathbb{R}^3$, yielding sparse scene-flow samples $\mathbf{S}_i$ for each generated view $I_i$.
Because the flow fields are estimated independently for each view, the resulting sparse scene-flow samples often exhibit directional and magnitude inconsistencies, even across geometrically corresponding regions (Fig.~\ref{fig:align}).
A related alignment objective has been explored in 4DGS-Cinemagraphy (3D-MOM)~\cite{jin2025optimizing}, which enforces cross-view consistency by optimizing reprojected 2D flows. However, since this approach relies on random initialization and image-space supervision, it may suffer from slow convergence and unstable alignment in non-overlapping or view-specific regions.

\setlength{\intextsep}{4pt} 
\begin{wrapfigure}{r}{0.45\linewidth}
  \centering
  % \vspace{-14pt}
  \includegraphics[width=\linewidth]{figure/align.pdf}
  
  \caption{Geometry-Aware Alignment Module for resolving local ambiguity and producing globally consistent scene-flow samples.}
  \label{fig:align}
  % \vspace{-13pt}
\end{wrapfigure}

To address these limitations, we introduce a geometry-aware alignment module that aligns lifted 3D scene-flow samples directly in 3D using spatial correspondences between the current and previously generated views.
We establish spatial correspondences by reprojecting 3D samples from the current view into previously generated views.
To enforce spatial consistency, we align the current scene-flow samples $\mathbf{S}_i$
with the previously accumulated scene-flow samples
\[
\mathbf{S}_{\text{prev}} = \{\mathbf{S}_1, \dots, \mathbf{S}_{i-1}\}.
\]
We formulate the alignment as estimating the optimal rotation 
$\mathbf{R} \in SO(3)$ and uniform scale $s$ that minimize the discrepancy between corresponding flow vectors:
\begin{equation}
\arg \min_{\mathbf{R}, s}
\sum_{k \in \mathcal{M}}
\left\|
\mathbf{S}_\text{prev}^{(k)}
-
s\mathbf{R}\mathbf{S}_i^{(k)}
\right\|^2,
\end{equation}
where $\mathcal{M}$ denotes the set of spatial correspondences between the current and previously generated scenes.
We solve this problem using the \textit{Kabsch algorithm}~\cite{kabsch1976solution}, 
which provides a closed-form solution for the optimal rotation via singular value decomposition (SVD). 
The scale parameter $s$ is estimated through a one-dimensional least-squares solution. 
Since the Kabsch algorithm has linear complexity with respect to the number of points, 
this yields a fast and stable alignment procedure.
To further reduce residual discrepancies, we perform a lightweight gradient-based refinement step. 
This alignment module consolidates independently estimated scene-flow samples into a globally consistent representation, providing stable supervision for the subsequent motion field learning stage.

\noindent\textbf{Motion Field Learning.}
The aligned scene-flow samples provide globally consistent motion supervision. 
However, they are only defined at discrete spatial locations and therefore cannot directly provide motion at newly queried positions during propagation. 
To support continuous motion queries throughout the scene, we instead learn a continuous global motion field.
Existing dynamic Gaussian methods typically follow a \emph{Lagrangian} formulation, where temporal dynamics are modeled by predicting deformation parameters for individual Gaussian primitives~\cite{wu20244d,bae2024per,li2024spacetime}. 
While expressive, this representation tightly couples motion modeling with the number of Gaussians, making optimization increasingly expensive as scene complexity grows.

In contrast, we adopt an \emph{Eulerian} formulation extended to 3D space.
Rather than estimating motion for each Gaussian independently, we learn a continuous global motion field
\(F_\theta : \mathbb{R}^3 \rightarrow \mathbb{R}^3\) that maps any spatial location \(\mathbf{x}\) to a velocity vector.
This representation decouples the motion estimation from individual primitives, allowing motion to be queried directly at arbitrary spatial locations.
As a result, Gaussian motion can be updated efficiently without per-Gaussian optimization, enabling scalable motion propagation across dynamically expanding scenes.
Learning such a continuous motion field in unstructured 3D space presents two challenges:
(1) 3D space lacks a regular grid structure, and  
(2) motion supervision is available only sparsely through estimated scene-flow samples.

To address these challenges, we parameterize the motion field using a multi-resolution hash encoding~\cite{muller2022instant} together with a lightweight MLP.
At each resolution level, the input position \( \mathbf{x} \in \mathbb{R}^3 \) is discretized and mapped into a compact hash table via spatial hashing:
\begin{equation}
h_\ell(\mathbf{x}) = \left( \bigoplus_{j=1}^{3} \lfloor x_j^\ell \rfloor \cdot \pi_j \right) \bmod T,
\end{equation}
where \( \pi_j \) are large primes, \( \bigoplus \) denotes bitwise XOR, and \( T \) is the size of the hash table (e.g., \( T = 2^{19} \)).  
This produces feature indices into learnable embedding tables across multiple resolutions. 
Features from all levels are interpolated and concatenated, and the aggregated feature vector is passed to an MLP to regress the motion:
\begin{equation}
F_\theta(\mathbf{x}) = \text{MLP}_\theta \left( \bigoplus_{\ell=1}^{L} \text{Interp}_\ell \left( \text{HashEnc}_\ell(\mathbf{x}) \right) \right).
\end{equation}

Given sparse scene-flow samples \( \{ (\mathbf{x}_i, \mathbf{s}_i) \}_{i=1}^N \), the model is trained with a simple regression objective:
\begin{equation}
\mathcal{L}_{\text{motion}} = \sum_{i=1}^N \left\| F_\theta(\mathbf{x}_i) - \mathbf{s}_i \right\|_2^2.
\end{equation}
The aligned scene flows obtained from the previous stage provide sparse yet globally consistent supervision for learning this field.
Once trained, the motion field predicts temporally coherent velocity vectors at arbitrary spatial locations, providing continuous supervision for Gaussian propagation during rendering.

\vspace{-2mm}
\subsection{Motion Propagation and Rendering}
\vspace{-0.5mm}
\label{motion_prop}
Although the learned motion field provides continuous velocity predictions, repeatedly advecting Gaussian primitives gradually causes them to drift away from their original spatial distribution. As a result, the rendered scene may develop density gaps in highly dynamic regions. Since our framework does not rely on iterative video-based refinement to recover these regions, we explicitly address this issue during motion propagation.

\noindent\textbf{Bidirectional Motion Propagation.}
To preserve spatial coverage during propagation, we advect Gaussians along the learned motion field in both forward and backward directions, forming a temporally symmetric motion cycle.
Similar to Eulerian flow integration in image space (Eqs.~\ref{eq:euler_motion}, \ref{eq:euler_integral}), 
we update the position of each Gaussian using discrete Euler integration over the learned motion field 
\(F_{\theta} : \mathbb{R}^3 \rightarrow \mathbb{R}^3\).

Given a Gaussian primitive \(g\) with center position \(\mathbf{p}_g(t) \in \mathbb{R}^3\) at time \(t\), we define its bidirectional trajectories using forward and backward integration:
\begin{equation}
\mathbf{p}_g^{f}(t)
=
\mathbf{p}_g^{f}(t-1)
+
\boldsymbol{\psi}\odot
F_{\theta}\!\big(\mathbf{p}_g^{f}(t-1)\big),
\label{eq:euler_forward}
\end{equation}
\vspace{-2mm}
\begin{equation}
\mathbf{p}_g^{b}(t)
=
\mathbf{p}_g^{b}(t-1)
-
\boldsymbol{\psi}\odot
F_{\theta}\!\big(\mathbf{p}_g^{b}(t-1)\big),
\label{eq:euler_backward}
\end{equation}
where \(\boldsymbol{\psi}\in\mathbb{R}^3\) denotes a per-axis step size vector, and \(\odot\) indicates element-wise multiplication.  
Forward integration advects Gaussians along the predicted velocity direction,
while backward integration propagates motion in the opposite direction.
This bidirectional formulation allows the motion trajectories to loop back toward the initial configuration,
helping maintain spatial coverage of the Gaussian distribution without introducing additional scene supervision.

After computing both trajectories, we construct the dynamic Gaussian set by combining the forward and backward trajectories:
\begin{equation}
{\small
\begin{aligned}
\mathcal{P}(t) =
\big\{ \mathbf{p}_g^{f}(t), \mathbf{p}_g^{b}(T - t) \mid \mathbf{m}_g = 1 \big\}
\cup
\big\{ \mathbf{p}_g^{\text{static}} \mid \mathbf{m}_g = 0 \big\}.
\end{aligned}
\label{eq:masked_concat}}
\end{equation}
Only Gaussians within motion regions (\(\mathbf{m}_g = 1\)) are updated and merged, while static ones remain fixed, reducing unnecessary rendering overhead.
To smoothly transition between the forward and backward trajectories, we modulate their opacities, $\alpha_g^{f}$ and $\alpha_g^{b}$, using a bidirectional opacity scheduler during rendering.
The contributions of the two trajectories are controlled by a time-dependent blending weight \(w(t)\in[0,1]\), where we use a simple linear schedule \(w(t)=t/T\):
\begin{equation}
\tilde{\alpha}_g^{f}(t) = \bigl(1 - w(t)\bigr)\,\alpha_g^{f},
\end{equation}
\vspace{-2mm}
\begin{equation}
\tilde{\alpha}_g^{b}(t) = w(t)\,\alpha_g^{b}.
\end{equation}
The opacity scheduler gradually shifts the dominant contribution from the forward trajectory to the backward trajectory, allowing the scene to smoothly return to its initial configuration while maintaining stable volumetric rendering. 
Together, bidirectional propagation and the opacity scheduler preserve spatial coverage of the Gaussian distribution while improving temporal stability during rendering. A complete summary of the iterative pipeline is provided in Algorithm~1 of the Supplementary Material \ref{sec: Algorithms_1}.

\vspace{-2mm}
\section{Experiments}
\vspace{-2mm}
\subsection{Baselines}
\vspace{-1mm}
In principle, we aim to compare our framework against existing state-of-the-art methods.
However, to the best of our knowledge, no prior work directly addresses the task of \textit{interactive 4D world generation} from a single image with controllable camera and motion hints.
Therefore, we organize the baselines into two groups that reflect the closest related research directions discussed in Sec.~\ref{sec:intro}.

\noindent\textbf{Video Generation Baselines.}
Recent controllable video generation models can synthesize temporally coherent videos conditioned on camera motion or control signals.
We include three representative methods from this line of work: \textit{Veo 3.1}~\cite{wiedemer2025video}, \textit{CogVideoX}~\cite{yang2024cogvideox} and \textit{Tora}~\cite{zhang2025tora}.
Although these models produce visually compelling videos, they do not construct an explicit 3D scene representation or motion field.
As a result, they cannot maintain geometrically consistent dynamics across viewpoints or support scene expansion.
Following prior work, we therefore evaluate them using non-reference metrics that assess perceptual quality and physical plausibility of the generated videos.

\noindent\textbf{4D Scene-based Baselines.}
We further compare against recent 4D scene generation methods with explicit dynamic scene representations:
\emph{4DGS-Cinemagraphy}~\cite{jin2025optimizing} and \emph{PerpetualWonder}~\cite{zhan2026perpetualwonder}.
4DGS-Cinemagraphy optimizes 4D Gaussians from a single landscape image and is closely related to environmental motion generation. PerpetualWonder combines generative modeling with physics-based simulation, providing a complementary comparison for physically driven dynamic scenes. Although these methods generate dynamic 4D content, they are not designed for interactive world expansion with controllable camera navigation and environmental motion hints. Thus, they serve as the closest 4D scene-based alternatives to our framework.

\vspace{-2.5mm}
\subsection{Evaluation Protocol}
\vspace{-0.5mm}
To quantitatively evaluate the generated 4D scenes, we assess both visual quality and motion quality. 
We construct a benchmark of 60 scenes from public images collected from Pexels~\cite{pexels} and Unsplash~\cite{unsplash}, covering four categories of environmental dynamics: clouds, water, smoke/fog, and fire, with 15 scenes per category. Following recent 4D scene generation works, we adopt VBench~\cite{huang2024vbench} metrics to evaluate rendered video quality. Specifically, \emph{Imaging Quality} and \emph{Aesthetic Quality} measure visual quality, while \emph{Motion Smoothness} and \emph{Temporal Flickering} evaluate temporal consistency and motion quality.
To assess the physical plausibility of generated environmental dynamics, we additionally employ the GPT-5.5-based \emph{PhysReal} metric following PhysGen3D~\cite{chen2025physgen3d}. PhysReal evaluates whether the generated environmental motion exhibits realistic physical behavior, complementing the perceptual and temporal quality metrics provided by VBench. All methods are evaluated under the same benchmark and protocol.

\vspace{-2mm}
\subsection{Implementation Details}
\vspace{-1mm}
We employ the Eulerian flow prediction model from 3D Cinemagraphy~\cite{li20233d} to estimate dense 2D motion fields.
Motion regions are extracted from user-provided control points using the Segment Anything Model (SAM)~\cite{kirillov2023segment}.
Depth is estimated using the pretrained MoGeV2 model~\cite{wang2026moge}.
All models are publicly available and used without additional fine-tuning to isolate the contribution of our motion field construction and propagation framework. Additional implementation details are provided in the Supplementary Material \ref{sec:impl_main}.
\vspace{-1mm}

\begin{table}[t]
\centering
\footnotesize
\setlength{\tabcolsep}{2.5pt}
\renewcommand{\arraystretch}{1.0}
\resizebox{\columnwidth}{!}{
\begin{tabular}{@{}l|l|c|c|c|c|c|c@{}}
\toprule
\textbf{Category} & \textbf{Method}
& \multicolumn{4}{c|}{\textbf{VBench (↑)}}
& \textbf{GPT-based (↑)}
& \textbf{Runtime (↓)} \\
\cmidrule(lr){3-6}
\cmidrule(lr){7-7}
\cmidrule(l){8-8}
&
& \textbf{Imaging}
& \textbf{Aesthetic}
& \textbf{Motion}
& \textbf{Flicker}
& \textbf{PhysReal}
& \textbf{Time (s)} \\
\midrule

\multirow{3}{*}{Video Gen.}
& Veo 3.1~\cite{wiedemer2025video}
& \textbf{0.694}
& \underline{0.625}
& 0.992
& 0.979
& \underline{0.622}
& \underline{140} \\

& CogVideoX~\cite{yang2024cogvideox}
& \underline{0.677}
& 0.611
& 0.991
& 0.983
& 0.575
& 1510 \\

& Tora~\cite{zhang2025tora}
& 0.649
& 0.609
& 0.992
& 0.976
& 0.571
& 550 \\

\midrule

\multirow{3}{*}{4D Scene}
& 4DGS-Cinemagraphy~\cite{jin2025optimizing}
& 0.637
& 0.604
& \textbf{0.996}
& \underline{0.988}
& 0.605
& 1980 \\

& PerpetualWonder~\cite{zhan2026perpetualwonder}
& 0.553
& 0.553
& 0.979
& 0.972
& 0.554
& 3580 \\

& \textbf{LivingWorld (Ours)}
& 0.673
& \textbf{0.639}
& \underline{0.995}
& \textbf{0.989}
& \textbf{0.655}
& \textbf{12} \\

\bottomrule
\end{tabular}
}
\vspace{1mm}
\caption{Comparison of environmental dynamics realism and visual quality across video generation and 4D scene-based methods. Best and second-best results are highlighted in bold and underlined, respectively.}
\label{tab:gpt}
\vspace{-6mm}
\end{table}

\begin{table}[t]
\centering
\footnotesize
\setlength{\tabcolsep}{5pt}
\renewcommand{\arraystretch}{1.05}
\resizebox{0.85\columnwidth}{!}{
\begin{tabular}{l|c|c|c|c}
\toprule
\textbf{Comparison} & \textbf{Imaging} & \textbf{Aesthetic} & \textbf{Motion} & \textbf{Flicker} \\
\midrule
Ours vs. Veo~3.1~\cite{wiedemer2025video}
& 58\%($\pm$9.5) & 66\%($\pm$9.1) & 72\%($\pm$8.7) & 75\%($\pm$8.4) \\
Ours vs. CogVideoX~\cite{yang2024cogvideox}
& 68\%($\pm$9.0) & 73\%($\pm$8.6) & 66\%($\pm$9.1) & 78\%($\pm$8.0) \\
Ours vs. Tora~\cite{zhang2025tora}
& 77\%($\pm$8.2) & 79\%($\pm$7.9) & 76\%($\pm$8.3) & 84\%($\pm$7.2) \\
Ours vs. 4DGS-Cinemagraphy~\cite{jin2025optimizing}
& 73\%($\pm$8.6) & 75\%($\pm$8.4) & 68\%($\pm$9.0) & 74\%($\pm$8.5) \\
Ours vs. PerpetualWonder~\cite{zhan2026perpetualwonder}
& 82\%($\pm$7.5) & 87\%($\pm$6.6) & 85\%($\pm$7.0) & 92\%($\pm$5.4) \\
\bottomrule
\end{tabular}
}
\vspace{1mm}
\caption{2AFC human preference study. Values indicate win rates of LivingWorld with 95\% confidence intervals.}
\label{tab:human_2afc}
\vspace{-3.5mm}
\end{table}

\vspace{-3mm}
\subsection{Quantitative Results}
\vspace{-1mm}

Table~\ref{tab:gpt} compares LivingWorld with recent video generation models and 4D scene generation approaches. Video generation models such as Veo~3.1~\cite{wiedemer2025video} achieve the strongest frame-wise visual quality, while LivingWorld remains visually competitive and achieves the best or near-best temporal quality across diverse environmental dynamics. LivingWorld also attains the highest PhysReal score, indicating that its generated dynamics are more physically plausible.
Compared with existing 4D scene generation methods, LivingWorld consistently outperforms 4DGS-Cinemagraphy~\cite{jin2025optimizing} and PerpetualWonder~\cite{zhan2026perpetualwonder} in overall quality. Moreover, LivingWorld generates a complete 4D scene in only 12 seconds, including approximately 9 seconds for scene generation and 3 seconds for motion alignment and motion field updates, whereas existing video generation and 4D scene generation approaches typically require several minutes.
To complement automatic metrics, we conduct a 2AFC human preference study in Table~\ref{tab:human_2afc}. LivingWorld is consistently preferred over all baselines, particularly on temporal criteria such as Motion and Flicker, further supporting the temporal coherence of the generated environmental dynamics. Details of the study protocol are provided in the Supplementary Material \ref{detail_2afc}.
These results suggest that explicitly modeling environmental dynamics as a global motion field provides a favorable balance between physical plausibility, temporal coherence, visual quality, and efficiency. Additional quantitative results and an interactivity user study evaluating the usability and convenience of our system are also provided in the Supplementary Material \ref{add_exp_1} and \ref{subsec:human_interactivity}.

\begin{figure*}[!t]
\centering
\includegraphics[width=\textwidth]{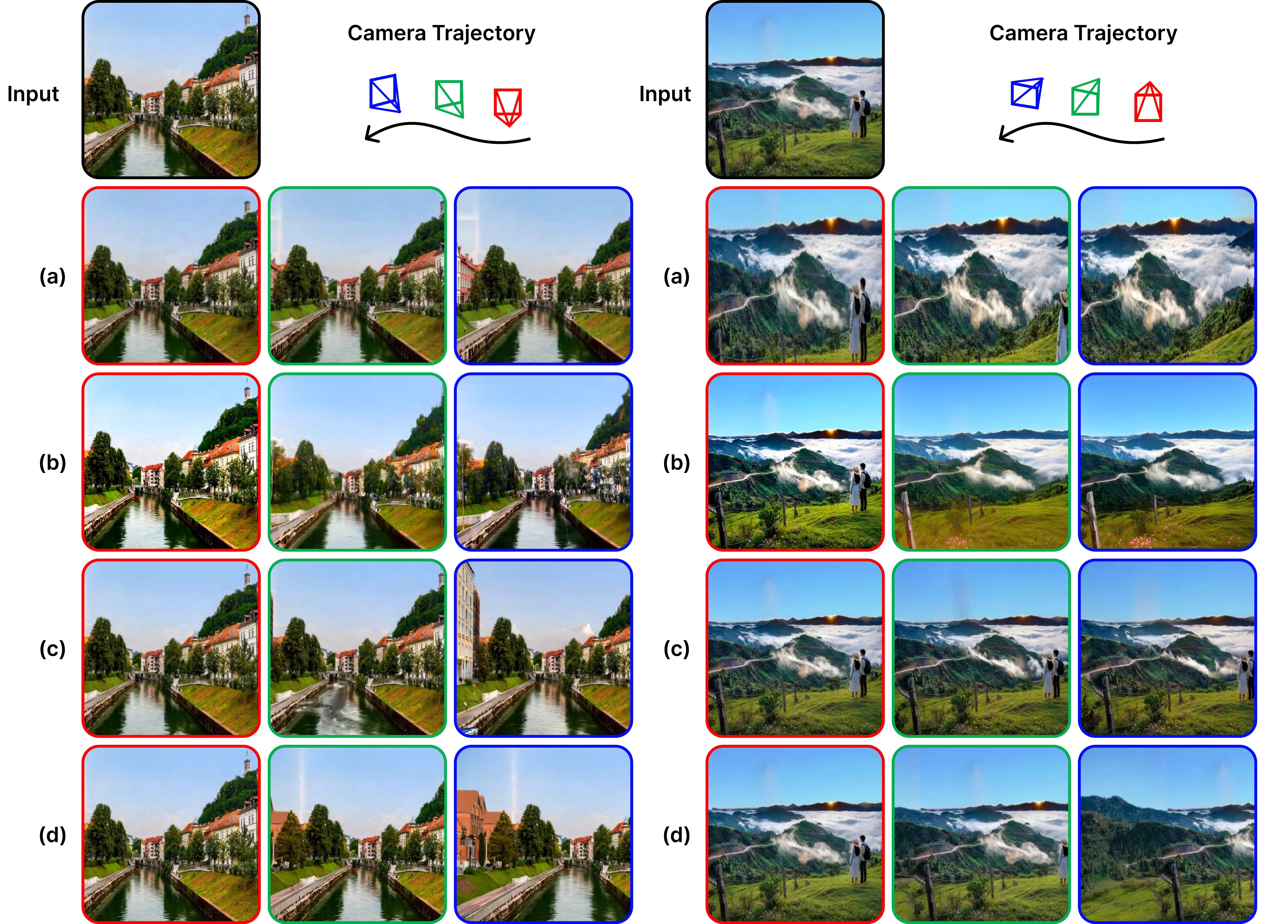}
\vspace{-5mm}
\caption{Qualitative comparison under camera movement. Each column corresponds to the camera viewpoint indicated by the colored camera icons. (a) Veo~3.1~\cite{wiedemer2025video}, (b) CogVideoX~\cite{yang2024cogvideox}, (c) Tora~\cite{zhang2025tora}, (d) 4DGS-Cinemagraphy~\cite{jin2025optimizing}, (e) PerpetualWonder~\cite{zhan2026perpetualwonder}, (f) Ours.
}
\label{fig:qual_baselines}
\vspace{-2mm}
\end{figure*}

\vspace{-2mm}
\subsection{Qualitative Results}
\vspace{-2mm}
Supplementary Material \ref{add_exp_4} provides additional qualitative results covering diverse environmental dynamics, including fluid-like motions such as fire, smoke, clouds, and water.
Fig.~\ref{fig:qual_baselines} presents a qualitative comparison with recent video generation and 4D scene generation methods. Veo~3.1~\cite{wiedemer2025video} often fails to generate meaningful environmental dynamics, leaving regions such as clouds or water largely static despite camera motion. CogVideoX~\cite{yang2024cogvideox} produces visually plausible motion but does not maintain consistent 3D scene structure, leading to noticeable geometric distortions and appearance inconsistencies as the camera viewpoint changes. Tora~\cite{zhang2025tora} exhibits temporal inconsistencies in dynamic regions, which produce visible artifacts in fluid-like areas. 4DGS-Cinemagraphy~\cite{jin2025optimizing} generates dynamic effects in selected regions, but the optimization-based scene flow often produces artifacts and unstable motion near dynamic boundaries, particularly in newly revealed areas where motion supervision is limited. PerpetualWonder~\cite{zhan2026perpetualwonder} is designed primarily for object-centric dynamics through physics-based simulation. While effective for localized object motion, it struggles to model diffuse environmental dynamics, which lack well-defined object boundaries and are spatially coupled with the surrounding scene. In contrast, our method (Fig.~\ref{fig:qual_baselines}f) preserves consistent scene geometry while generating coherent environmental motion, enabling stable dynamics across expanded views under controlled camera trajectories.

\begin{figure*}[!t]
\centering
\includegraphics[width=\textwidth]{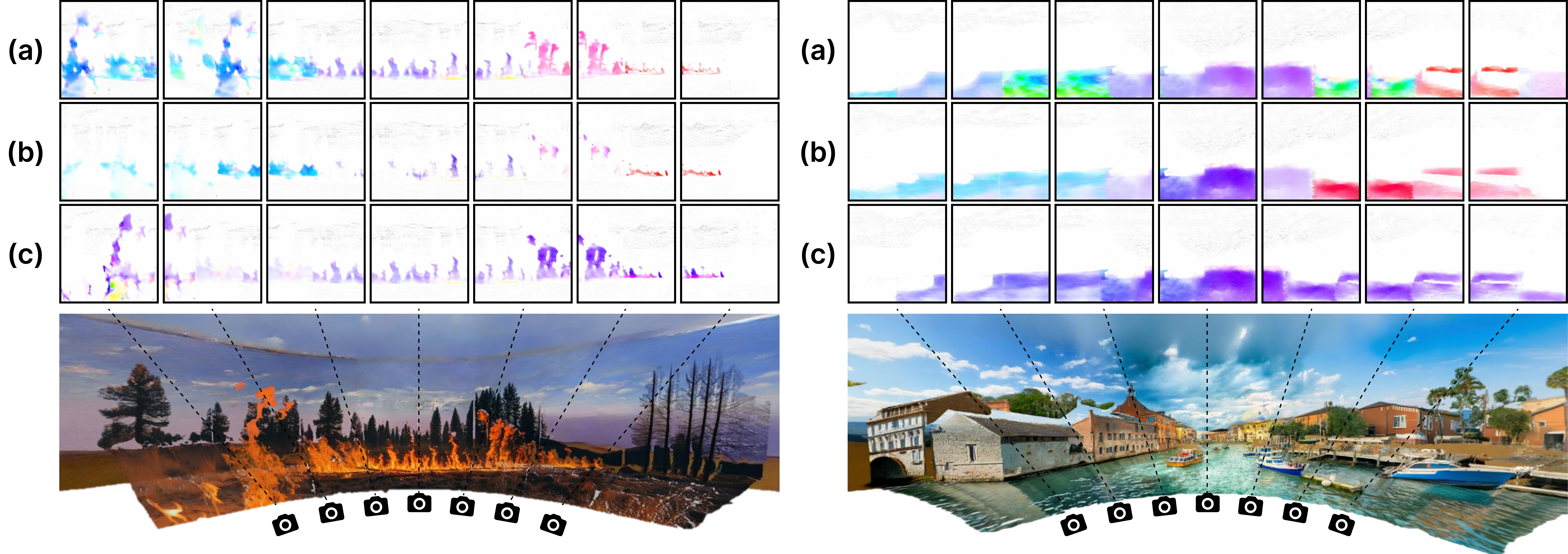}
\caption{Qualitative comparison of motion consistency across different methods on the same input image. (a) Naive Scene Flow, (b) WonderWorld~\cite{yu2025wonderworld} + 3D-MOM, and (c) Ours.}
\label{fig:ablation_qual}
\vspace{-2mm}
\end{figure*}

\begin{table}[t]
\centering
\footnotesize
\setlength{\tabcolsep}{4pt}
\renewcommand{\arraystretch}{1.05}
\resizebox{\columnwidth}{!}{
\begin{tabular}{l|c|cc|cc}
\toprule
\textbf{Method}
& \textbf{Runtime (s) ($\downarrow$)}
& \multicolumn{2}{c|}{\textbf{Local Consistency}}
& \multicolumn{2}{c}{\textbf{Global Consistency}} \\
\cmidrule(lr){3-4}
\cmidrule(l){5-6}
&
& \textbf{MCA ($\uparrow$)}
& \textbf{FMV ($\downarrow$)}
& \textbf{Cosine ($\uparrow$)}
& \textbf{Mag. Ratio ($\uparrow$)} \\
\midrule

WonderWorld~\cite{yu2025wonderworld} + Naive Scene Flow
& \textbf{9.3}
& 0.0550
& 1.91
& 0.54
& 0.72 \\

WonderWorld~\cite{yu2025wonderworld} + 3D-MOM~\cite{jin2025optimizing}
& 600.0
& 0.0597
& 1.66
& 0.66
& 0.76 \\

\textbf{Ours}
& 12.1
& \textbf{0.0742}
& \textbf{0.29}
& \textbf{0.91}
& \textbf{0.84} \\

\bottomrule
\end{tabular}
}
\vspace{1mm}
\caption{Quantitative comparison for the geometry-aware alignment module.
We compare naive scene flow propagation, optimization-based alignment (3D-MOM), and our geometry-aware alignment.}
\label{tab:time_mca_cost}
\vspace{-2.5mm}
\end{table}

\vspace{-4mm}
\subsection{Ablation study}
\vspace{-1.5mm}
We analyze the contribution of our key components: 
(1) the motion alignment module and 
(2) the design of the motion field, including the hash-based representation and bidirectional motion integration. Additional ablation studies on opacity blending schedules and boundary hole reduction are provided in the Supplementary Material \ref{supp:opacity} and \ref{supp:boundary}.

\noindent\textbf{Geometry-Aware Alignment Module.}
To evaluate the proposed motion alignment module, we compare three variants:
(1) naive scene flow accumulation without alignment,
(2) optimization-based alignment using 3D-MOM from 4DGS-Cinemagraphy~\cite{jin2025optimizing}, and
(3) our lightweight geometry-aware alignment module.
We evaluate local consistency using MCA and FMV, and global consistency using cosine similarity and magnitude ratio over spatially distant point pairs, including newly expanded areas. Detailed metric definitions are provided in the Supplementary Material \ref{Metrics}.
Table~\ref{tab:time_mca_cost} reports runtime, local consistency, and global consistency comparisons.
Naive accumulation produces unstable directions and inconsistent magnitudes, yielding the lowest global consistency.
3D-MOM improves local coherence through reprojection-based optimization, but it is restricted to overlapping regions and incurs high computational cost.
In contrast, our alignment module achieves the best local and global consistency with efficient runtime, improving MCA, reducing FMV, and achieving the highest cosine similarity and magnitude ratio.
Fig.~\ref{fig:ablation_qual} provides qualitative comparisons.
Naive scene flow (a) produces unstable motion with exaggerated magnitude accumulation.
3D-MOM (b) improves local coherence in overlapping regions but fails to maintain globally consistent motion in newly generated areas.
Our method (c) generates stable and coherent environmental dynamics across viewpoints, demonstrating the effectiveness of lightweight global motion alignment.

\begin{figure*}[!t]
\centering
\includegraphics[width=\textwidth]{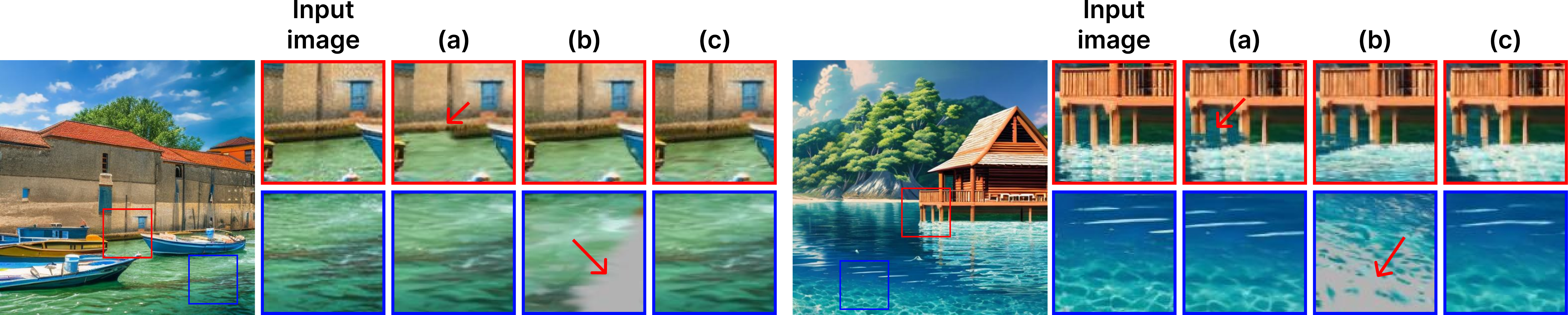}
\vspace{-5mm}
\caption{Ablation study on motion field design.
(a) w/o hash-based motion field, (b) w/o bidirectional motion propagation, (c) Ours }
\label{fig:ablation_motionfield}
\vspace{-1mm}
\end{figure*}

\vspace{-0.2mm}
\noindent\textbf{Motion Field Design.}
We analyze the impact of two key components in our framework: the global motion field and bidirectional motion propagation (Fig.~\ref{fig:ablation_motionfield}).
First, we evaluate a variant that directly applies the estimated 3D scene flow to Gaussians without learning a motion field (Fig.~\ref{fig:ablation_motionfield}a).
In this setting, motion is propagated through Euler integration using the same flow vector at each step.
As a result, motion vectors are repeatedly accumulated over time for each Gaussian, leading to unrealistic motion magnitudes and severe temporal artifacts, particularly in regions with complex environmental dynamics.
Next, we ablate bidirectional motion propagation by using only forward motion integration (Fig.~\ref{fig:ablation_motionfield}b).
Although this variant produces locally smooth motion in visible regions, it fails to preserve Gaussian density in highly dynamic areas.
Forward-only integration causes Gaussians to gradually drift away from their original distribution, resulting in visible holes and unstable rendering near motion boundaries.
In contrast, our full model (Fig.~\ref{fig:ablation_motionfield}c) learns a compact hash-based motion field that defines motion over the global 3D space.
Each Gaussian retrieves motion based on its spatial location, preventing redundant motion accumulation and improving temporal stability.
Combined with bidirectional propagation, our method maintains density in dynamic regions and produces stable, artifact-free environmental dynamics.

\vspace{-2mm}
\subsection{Integration with Object-Centric Motion}
\begin{figure*}[!t]
\centering
\includegraphics[width=\textwidth]{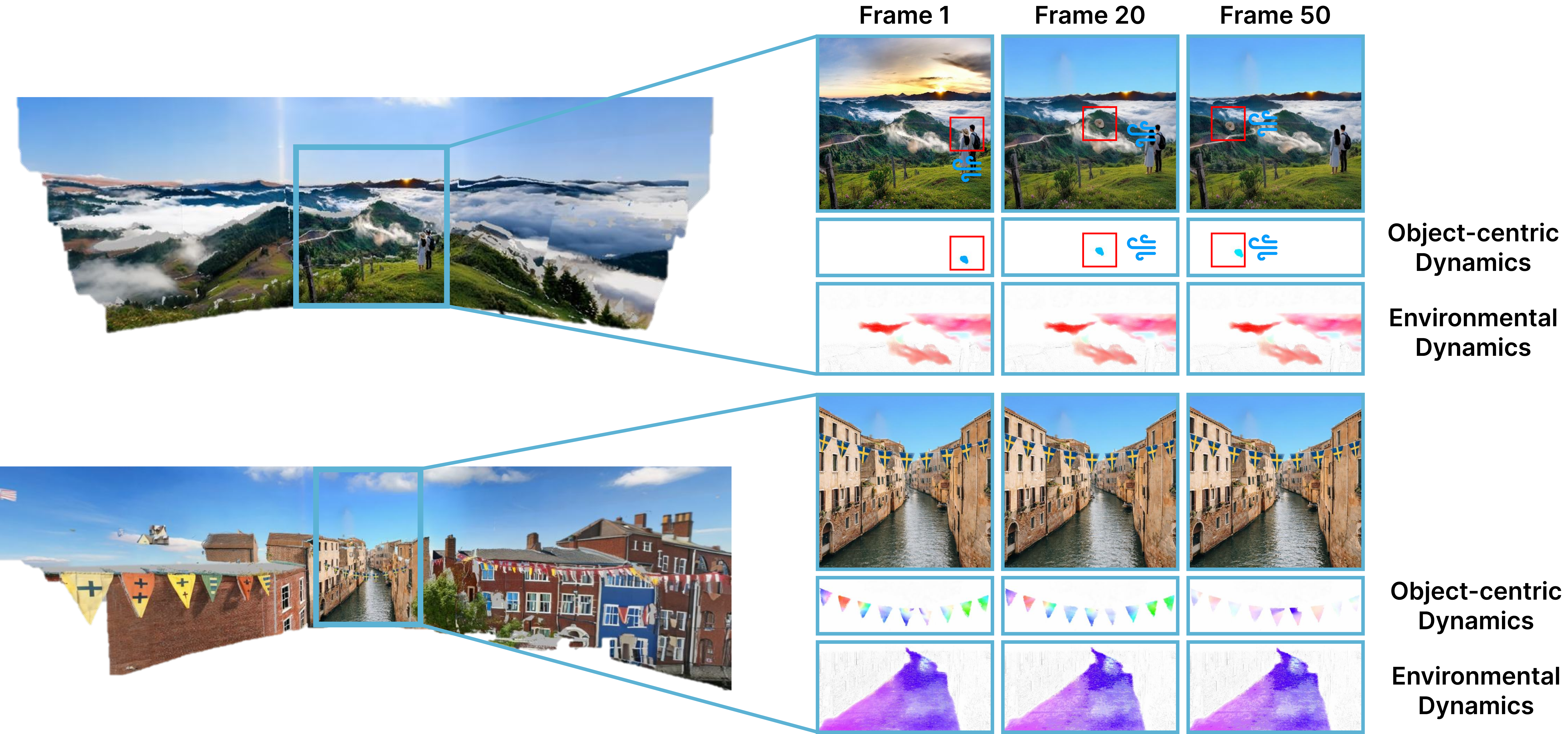}
\caption{Integration of environmental dynamics with localized object-centric motion. Environmental dynamics (e.g., water flow) coexist with motion applied to selected regions such as flags and tree branches.}
\label{fig:rigid}
\end{figure*}

\vspace{-1mm}
While our primary focus is modeling environmental dynamics such as clouds, water, and other fluid-like phenomena, our framework can also incorporate localized object-centric motion. 
To demonstrate this capability, we augment the environmental motion field with localized motion applied to selected regions, such as the fluttering of flags or the swaying of tree branches. 
As shown in Fig.~\ref{fig:rigid}, these object-centric motions integrate naturally with the generated environmental dynamics, producing visually consistent scene evolution. 
This example illustrates that the proposed framework can incorporate localized rigid motion while preserving its primary focus on large-scale environmental dynamics. 
In future work, the framework could be extended to support richer forms of object-centric motion, including articulated objects and human motion, enabling more complex interactive dynamic worlds.

\vspace{-3mm}

\vspace{-1mm}
\section{Conclusion}
\vspace{-2.5mm}
We presented a framework for interactive 4D world generation that explicitly models environmental dynamics within a reconstructed scene, enabling coherent motion propagation and efficient visual feedback during scene expansion. 
While our focus is environmental motion such as clouds and water, we demonstrate that the framework can also incorporate localized rigid motion, suggesting the potential to support a broader range of dynamic behaviors. 
Extending the framework to handle more complex object-centric motion, such as articulated objects or human motion, remains an important direction for future work, particularly in addressing challenges related to world-coordinate alignment and motion interactions.

\vspace{-3mm}
\section*{Acknowledgements}
\vspace{-2mm}
This work was supported by the National Research Foundation of Korea (NRF) grants funded by the Korean government (MSIT) (RS-2024-00456152). Computational resources were provided by “the Advanced GPU Utilization Support Program” funded by the Government of the Republic of Korea (Ministry of Science and ICT) and the Cluster Server for Computational Science at Pusan National University.

\newpage

\bibliographystyle{splncs04}
\bibliography{main}

@String(AAAI  = {AAAI})

@String(TOG   = {ACM Trans. Graph.})

@String(TOG   = {ACM TOG})

@incollection{chuang2005animating,
  title     = {Animating Pictures with Stochastic Motion Textures},
  author    = {Chuang, Yung-Yu and Goldman, Dan B. and Zheng, Ke Colin and Curless, Brian and Salesin, David H. and Szeliski, Richard},
  booktitle = {ACM SIGGRAPH 2005 Papers},
  pages     = {853--860},
  year      = {2005},
  publisher = {Association for Computing Machinery},
  doi       = {10.1145/1186822.1073273}
}

@article{jhou2015animating,
  title={Animating still landscape photographs through cloud motion creation},
  author={Jhou, Wei-Cih and Cheng, Wen-Huang},
  journal={IEEE Transactions on Multimedia},
  volume={18},
  number={1},
  pages={4--13},
  year={2015},
  publisher={IEEE}
}

@inproceedings{holynski2021animating,
  title={Animating pictures with eulerian motion fields},
  author={Holynski, Aleksander and Curless, Brian L and Seitz, Steven M and Szeliski, Richard},
  booktitle={Proceedings of the IEEE/CVF Conference on Computer Vision and Pattern Recognition},
  pages={5810--5819},
  year={2021}
}

@inproceedings{li20233d,
  title={3d cinemagraphy from a single image},
  author={Li, Xingyi and Cao, Zhiguo and Sun, Huiqiang and Zhang, Jianming and Xian, Ke and Lin, Guosheng},
  booktitle={Proceedings of the IEEE/CVF Conference on Computer Vision and Pattern Recognition},
  pages={4595--4605},
  year={2023}
}

@article{mahapatra2023text,
  title={Text-guided synthesis of eulerian cinemagraphs},
  author={Mahapatra, Aniruddha and Siarohin, Aliaksandr and Lee, Hsin-Ying and Tulyakov, Sergey and Zhu, Jun-Yan},
  journal={ACM Transactions on Graphics (TOG)},
  volume={42},
  number={6},
  pages={1--13},
  year={2023},
  publisher={ACM New York, NY, USA}
}

@inproceedings{jin2025optimizing,
  title={Optimizing 4D Gaussians for dynamic scene video from single landscape images},
  author={Jin, In-Hwan and Choo, Haesoo and Jeong, Seong-Hun and Heemoon, Park and Kim, Junghwan and Kwon, Oh-joon and Kong, Kyeongbo},
  booktitle={The Thirteenth International Conference on Learning Representations},
  year={2025}
}

@article{kerbl20233d,
  title={3D Gaussian splatting for real-time radiance field rendering.},
  author={Kerbl, Bernhard and Kopanas, Georgios and Leimk{\"u}hler, Thomas and Drettakis, George},
  journal={ACM Trans. Graph.},
  volume={42},
  number={4},
  pages={139--1},
  year={2023}
}

@article{mildenhall2021nerf,
  title={Nerf: Representing scenes as neural radiance fields for view synthesis},
  author={Mildenhall, Ben and Srinivasan, Pratul P and Tancik, Matthew and Barron, Jonathan T and Ramamoorthi, Ravi and Ng, Ren},
  journal={Communications of the ACM},
  volume={65},
  number={1},
  pages={99--106},
  year={2021},
  publisher={ACM New York, NY, USA}
}

@inproceedings{wu20244d,
  title={4d gaussian splatting for real-time dynamic scene rendering},
  author={Wu, Guanjun and Yi, Taoran and Fang, Jiemin and Xie, Lingxi and Zhang, Xiaopeng and Wei, Wei and Liu, Wenyu and Tian, Qi and Wang, Xinggang},
  booktitle={Proceedings of the IEEE/CVF conference on computer vision and pattern recognition},
  pages={20310--20320},
  year={2024}
}

@inproceedings{choi2024stylecinegan,
  title={StyleCineGAN: Landscape Cinemagraph Generation using a Pre-trained StyleGAN},
  author={Choi, Jongwoo and Seo, Kwanggyoon and Ashtari, Amirsaman and Noh, Junyong},
  booktitle={Proceedings of the IEEE/CVF Conference on Computer Vision and Pattern Recognition},
  pages={7872--7881},
  year={2024}
}

@inproceedings{shen2023make,
  title={Make-it-4d: Synthesizing a consistent long-term dynamic scene video from a single image},
  author={Shen, Liao and Li, Xingyi and Sun, Huiqiang and Peng, Juewen and Xian, Ke and Cao, Zhiguo and Lin, Guosheng},
  booktitle={Proceedings of the 31st ACM International Conference on Multimedia},
  pages={8167--8175},
  year={2023}
}

@inproceedings{yu2025wonderworld,
  title={Wonderworld: Interactive 3d scene generation from a single image},
  author={Yu, Hong-Xing and Duan, Haoyi and Herrmann, Charles and Freeman, William T and Wu, Jiajun},
  booktitle={Proceedings of the Computer Vision and Pattern Recognition Conference},
  pages={5916--5926},
  year={2025}
}

@article{kabsch1976solution,
  title={A solution for the best rotation to relate two sets of vectors},
  author={Kabsch, Wolfgang},
  journal={Foundations of Crystallography},
  volume={32},
  number={5},
  pages={922--923},
  year={1976},
  publisher={International Union of Crystallography}
}

@inproceedings{bae2024per,
  title={Per-gaussian embedding-based deformation for deformable 3d gaussian splatting},
  author={Bae, Jeongmin and Kim, Seoha and Yun, Youngsik and Lee, Hahyun and Bang, Gun and Uh, Youngjung},
  booktitle={European Conference on Computer Vision},
  pages={321--335},
  year={2024},
  organization={Springer}
}

@inproceedings{li2024spacetime,
  title={Spacetime gaussian feature splatting for real-time dynamic view synthesis},
  author={Li, Zhan and Chen, Zhang and Li, Zhong and Xu, Yi},
  booktitle={Proceedings of the IEEE/CVF Conference on Computer Vision and Pattern Recognition},
  pages={8508--8520},
  year={2024}
}

@article{muller2022instant,
  title={Instant neural graphics primitives with a multiresolution hash encoding},
  author={M{\"u}ller, Thomas and Evans, Alex and Schied, Christoph and Keller, Alexander},
  journal={ACM transactions on graphics (TOG)},
  volume={41},
  number={4},
  pages={1--15},
  year={2022},
  publisher={ACM New York, NY, USA}
}

@article{kaneva2010infinite,
  title={Infinite images: Creating and exploring a large photorealistic virtual space},
  author={Kaneva, Biliana and Sivic, Josef and Torralba, Antonio and Avidan, Shai and Freeman, William T},
  journal={Proceedings of the IEEE},
  volume={98},
  number={8},
  pages={1391--1407},
  year={2010},
  publisher={IEEE}
}

@inproceedings{liu2021infinite,
  title={Infinite nature: Perpetual view generation of natural scenes from a single image},
  author={Liu, Andrew and Tucker, Richard and Jampani, Varun and Makadia, Ameesh and Snavely, Noah and Kanazawa, Angjoo},
  booktitle={Proceedings of the IEEE/CVF International Conference on Computer Vision},
  pages={14458--14467},
  year={2021}
}

@article{wu2024blockfusion,
  title={Blockfusion: Expandable 3d scene generation using latent tri-plane extrapolation},
  author={Wu, Zhennan and Li, Yang and Yan, Han and Shang, Taizhang and Sun, Weixuan and Wang, Senbo and Cui, Ruikai and Liu, Weizhe and Sato, Hiroyuki and Li, Hongdong and others},
  journal={ACM Transactions on Graphics (ToG)},
  volume={43},
  number={4},
  pages={1--17},
  year={2024},
  publisher={ACM New York, NY, USA}
}

@inproceedings{hua2025sat2city,
  title={Sat2city: 3d city generation from a single satellite image with cascaded latent diffusion},
  author={Hua, Tongyan and Jiang, Lutao and Chen, Ying-Cong and Zhao, Wufan},
  booktitle={Proceedings of the IEEE/CVF International Conference on Computer Vision},
  pages={27978--27988},
  year={2025}
}

@inproceedings{zhou2025scenex,
  title     = {SceneX: Procedural Controllable Large-Scale Scene Generation},
  author    = {Zhou, Mengqi and Wang, Yuxi and Hou, Jun and Zhang, Shougao and Li, Yiwei and Luo, Chuanchen and Peng, Junran and Zhang, Zhaoxiang},
  booktitle = {Proceedings of the AAAI Conference on Artificial Intelligence},
  volume    = {39},
  pages     = {10806--10814},
  year      = {2025}
}

@inproceedings{yu2024wonderjourney,
  title={Wonderjourney: Going from anywhere to everywhere},
  author={Yu, Hong-Xing and Duan, Haoyi and Hur, Junhwa and Sargent, Kyle and Rubinstein, Michael and Freeman, William T and Cole, Forrester and Sun, Deqing and Snavely, Noah and Wu, Jiajun and others},
  booktitle={Proceedings of the IEEE/CVF Conference on Computer Vision and Pattern Recognition},
  pages={6658--6667},
  year={2024}
}

@inproceedings{shi2024motion,
  title={Motion-i2v: Consistent and controllable image-to-video generation with explicit motion modeling},
  author={Shi, Xiaoyu and Huang, Zhaoyang and Wang, Fu-Yun and Bian, Weikang and Li, Dasong and Zhang, Yi and Zhang, Manyuan and Cheung, Ka Chun and See, Simon and Qin, Hongwei and others},
  booktitle={ACM SIGGRAPH 2024 Conference Papers},
  pages={1--11},
  year={2024}
}

@inproceedings{xing2025motioncanvas,
  title={Motioncanvas: Cinematic shot design with controllable image-to-video generation},
  author={Xing, Jinbo and Mai, Long and Ham, Cusuh and Huang, Jiahui and Mahapatra, Aniruddha and Fu, Chi-Wing and Wong, Tien-Tsin and Liu, Feng},
  booktitle={Proceedings of the Special Interest Group on Computer Graphics and Interactive Techniques Conference Conference Papers},
  pages={1--11},
  year={2025}
}

@inproceedings{shi2025motionstone,
  title={Motionstone: Decoupled motion intensity modulation with diffusion transformer for image-to-video generation},
  author={Shi, Shuwei and Gong, Biao and Chen, Xi and Zheng, Dandan and Tan, Shuai and Yang, Zizheng and Li, Yuyuan and He, Jingwen and Zheng, Kecheng and Chen, Jingdong and others},
  booktitle={Proceedings of the Computer Vision and Pattern Recognition Conference},
  pages={22864--22874},
  year={2025}
}

@inproceedings{chen2025physgen3d,
  title={Physgen3d: Crafting a miniature interactive world from a single image},
  author={Chen, Boyuan and Jiang, Hanxiao and Liu, Shaowei and Gupta, Saurabh and Li, Yunzhu and Zhao, Hao and Wang, Shenlong},
  booktitle={Proceedings of the Computer Vision and Pattern Recognition Conference},
  pages={6178--6189},
  year={2025}
}

@article{singer2023text,
  title     = {Text-to-4D Dynamic Scene Generation},
  author    = {Singer, Uriel and Sheynin, Shelly and Polyak, Adam and Ashual, Oron and Makarov, Iurii and Kokkinos, Filippos and Goyal, Naman and Vedaldi, Andrea and Parikh, Devi and Johnson, Justin and others},
  journal   = {arXiv preprint arXiv:2301.11280},
  year      = {2023}
}

@inproceedings{bahmani20244d,
  title     = {4D-fy: Text-to-4D Generation using Hybrid Score Distillation Sampling},
  author    = {Bahmani, Sherwin and Skorokhodov, Ivan and Rong, Victor and Wetzstein, Gordon and Guibas, Leonidas and Wonka, Peter and Tulyakov, Sergey and Park, Jeong Joon and Tagliasacchi, Andrea and Lindell, David B},
  booktitle = {Proceedings of the IEEE/CVF Conference on Computer Vision and Pattern Recognition},
  pages     = {7996--8006},
  year      = {2024}
}

@inproceedings{zheng2024unified,
  title     = {A Unified Approach for Text- and Image-guided 4D Scene Generation},
  author    = {Zheng, Yufeng and Li, Xueting and Nagano, Koki and Liu, Sifei and Hilliges, Otmar and De Mello, Shalini},
  booktitle = {Proceedings of the IEEE/CVF Conference on Computer Vision and Pattern Recognition},
  pages     = {7300--7309},
  year      = {2024}
}

@inproceedings{bahmani2024tc4d,
  title     = {TC4D: Trajectory-Conditioned Text-to-4D Generation},
  author    = {Bahmani, Sherwin and Liu, Xian and Yifan, Wang and Skorokhodov, Ivan and Rong, Victor and Liu, Ziwei and Liu, Xihui and Park, Jeong Joon and Tulyakov, Sergey and Wetzstein, Gordon and others},
  booktitle = {European Conference on Computer Vision},
  pages     = {53--72},
  year      = {2024},
  organization = {Springer}
}

@article{zeng2024trans4d,
  title     = {Trans4D: Realistic Geometry-aware Transition for Compositional Text-to-4D Synthesis},
  author    = {Zeng, Bohan and Yang, Ling and Li, Siyu and Liu, Jiaming and Zhang, Zixiang and Tian, Juanxi and Zhu, Kaixin and Guo, Yongzhen and Wang, Fu-Yun and Xu, Minkai and others},
  journal   = {arXiv preprint arXiv:2410.07155},
  year      = {2024}
}

@inproceedings{wu2025cat4d,
  title     = {CAT4D: Create Anything in 4D with Multi-View Video Diffusion Models},
  author    = {Wu, Rundi and Gao, Ruiqi and Poole, Ben and Trevithick, Alex and Zheng, Changxi and Barron, Jonathan T. and Holynski, Aleksander},
  booktitle = {Proceedings of the IEEE/CVF Conference on Computer Vision and Pattern Recognition},
  pages     = {26057--26068},
  year      = {2025}
}

@inproceedings{li2025wonderplay,
  title={Wonderplay: Dynamic 3d scene generation from a single image and actions},
  author={Li, Zizhang and Yu, Hong-Xing and Liu, Wei and Yang, Yin and Herrmann, Charles and Wetzstein, Gordon and Wu, Jiajun},
  booktitle={Proceedings of the IEEE/CVF International Conference on Computer Vision},
  pages={9080--9090},
  year={2025}
}

@inproceedings{kirillov2023segment,
  title={Segment anything},
  author={Kirillov, Alexander and Mintun, Eric and Ravi, Nikhila and Mao, Hanzi and Rolland, Chloe and Gustafson, Laura and Xiao, Tete and Whitehead, Spencer and Berg, Alexander C and Lo, Wan-Yen and others},
  booktitle={Proceedings of the IEEE/CVF international conference on computer vision},
  pages={4015--4026},
  year={2023}
}

@inproceedings{hollein2023text2room,
  title={Text2room: Extracting textured 3d meshes from 2d text-to-image models},
  author={H{\"o}llein, Lukas and Cao, Ang and Owens, Andrew and Johnson, Justin and Nie{\ss}ner, Matthias},
  booktitle={Proceedings of the IEEE/CVF International Conference on Computer Vision},
  pages={7909--7920},
  year={2023}
}

@article{chung2023luciddreamer,
  title={Luciddreamer: Domain-free generation of 3d gaussian splatting scenes},
  author={Chung, Jaeyoung and Lee, Suyoung and Nam, Hyeongjin and Lee, Jaerin and Lee, Kyoung Mu},
  journal={arXiv preprint arXiv:2311.13384},
  year={2023}
}

@article{ouyang2023text2immersion,
  title={Text2immersion: Generative immersive scene with 3d gaussians},
  author={Ouyang, Hao and Heal, Kathryn and Lombardi, Stephen and Sun, Tiancheng},
  journal={arXiv preprint arXiv:2312.09242},
  year={2023}
}

@article{endo2019animating,
  title={Animating landscape: self-supervised learning of decoupled motion and appearance for single-image video synthesis},
  author={Endo, Yuki and Kanamori, Yoshihiro and Kuriyama, Shigeru},
  journal={arXiv preprint arXiv:1910.07192},
  year={2019}
}

@inproceedings{logacheva2020deeplandscape,
  title={Deeplandscape: Adversarial modeling of landscape videos},
  author={Logacheva, Elizaveta and Suvorov, Roman and Khomenko, Oleg and Mashikhin, Anton and Lempitsky, Victor},
  booktitle={European Conference on Computer Vision},
  pages={256--272},
  year={2020},
  organization={Springer}
}

@article{wang2026moge,
  title={Moge-2: Accurate monocular geometry with metric scale and sharp details},
  author={Wang, Ruicheng and Xu, Sicheng and Dong, Yue and Deng, Yu and Xiang, Jianfeng and Lv, Zelong and Sun, Guangzhong and Tong, Xin and Yang, Jiaolong},
  journal={Advances in Neural Information Processing Systems},
  volume={38},
  pages={35928--35959},
  year={2026}
}

@article{zhan2026perpetualwonder,
  title={PerpetualWonder: Long-Horizon Action-Conditioned 4D Scene Generation},
  author={Zhan, Jiahao and Li, Zizhang and Yu, Hong-Xing and Wu, Jiajun},
  journal={arXiv preprint arXiv:2602.04876},
  year={2026}
}

@inproceedings{zhang2025tora,
  title={Tora: Trajectory-oriented diffusion transformer for video generation},
  author={Zhang, Zhenghao and Liao, Junchao and Li, Menghao and Dai, Zuozhuo and Qiu, Bingxue and Zhu, Siyu and Qin, Long and Wang, Weizhi},
  booktitle={Proceedings of the Computer Vision and Pattern Recognition Conference},
  pages={2063--2073},
  year={2025}
}

@article{yang2024cogvideox,
  title={Cogvideox: Text-to-video diffusion models with an expert transformer},
  author={Yang, Zhuoyi and Teng, Jiayan and Zheng, Wendi and Ding, Ming and Huang, Shiyu and Xu, Jiazheng and Yang, Yuanming and Hong, Wenyi and Zhang, Xiaohan and Feng, Guanyu and others},
  journal={arXiv preprint arXiv:2408.06072},
  year={2024}
}

@article{wiedemer2025video,
  title={Video models are zero-shot learners and reasoners},
  author={Wiedemer, Thadd{\"a}us and Li, Yuxuan and Vicol, Paul and Gu, Shixiang Shane and Matarese, Nick and Swersky, Kevin and Kim, Been and Jaini, Priyank and Geirhos, Robert},
  journal={arXiv preprint arXiv:2509.20328},
  year={2025}
}

@article{baldridge2024imagen,
  title={Imagen 3},
  author={Baldridge, Jason and Bauer, Jakob and Bhutani, Mukul and Brichtova, Nicole and Bunner, Andrew and Castrejon, Lluis and Chan, Kelvin and Chen, Yichang and Dieleman, Sander and Du, Yuqing and others},
  journal={arXiv preprint arXiv:2408.07009},
  year={2024}
}

@article{wan2025wan,
  title={Wan: Open and advanced large-scale video generative models},
  author={Wan, Team and Wang, Ang and Ai, Baole and Wen, Bin and Mao, Chaojie and Xie, Chen-Wei and Chen, Di and Yu, Feiwu and Zhao, Haiming and Yang, Jianxiao and others},
  journal={arXiv preprint arXiv:2503.20314},
  year={2025}
}

@inproceedings{huang2024vbench,
  title={Vbench: Comprehensive benchmark suite for video generative models},
  author={Huang, Ziqi and He, Yinan and Yu, Jiashuo and Zhang, Fan and Si, Chenyang and Jiang, Yuming and Zhang, Yuanhan and Wu, Tianxing and Jin, Qingyang and Chanpaisit, Nattapol and others},
  booktitle={Proceedings of the IEEE/CVF Conference on Computer Vision and Pattern Recognition},
  pages={21807--21818},
  year={2024}
}

@misc{pexels,
  author = {{Pexels}},
  title = {Royalty-Free Stock Footage Website},
  howpublished = {\url{https://www.pexels.com}},
  note = {Accessed: 2026-06-28}
}

@misc{unsplash,
  author       = {{Unsplash}},
  title        = {Unsplash: Beautiful Free Images and Pictures},
  year         = {2026},
  howpublished = {\url{https://unsplash.com}},
  note         = {Accessed: June 2026}
}

\newpage

\appendix

% Fix duplicated hyperref anchors between main paper and appendix
\makeatletter
\renewcommand{\theHsection}{supp.\Alph{section}}
\renewcommand{\theHsubsection}{supp.\Alph{section}.\arabic{subsection}}
\renewcommand{\theHsubsubsection}{supp.\Alph{section}.\arabic{subsection}.\arabic{subsubsection}}
\renewcommand{\theHfigure}{supp.\Alph{section}.\arabic{figure}}
\renewcommand{\theHtable}{supp.\Alph{section}.\arabic{table}}
\renewcommand{\theHequation}{supp.\Alph{section}.\arabic{equation}}
\makeatother
\section*{Supplementary Material Overview}
This supplementary material provides additional algorithmic details,
implementation details, motion consistency metrics, ablation analyses,
human study protocols, and experimental results that support the main paper.

\begin{itemize}

\item \textbf{A. Algorithmic Formulation}
\begin{itemize}
    \item \ref{sec: Algorithms_1} LivingWorld Interactive Pipeline
\end{itemize}

\item \textbf{B. Implementation Details}
\begin{itemize}
    \item \ref{supp:impl_correspondence} Reprojection-based Correspondence Construction
    \item \ref{supp:impl_alignment} Alignment Initialization and Refinement
    \item \ref{supp:impl_hash} Hash-based Motion Field
\end{itemize}

\item \textbf{C. Motion Consistency Metrics}
\begin{itemize}
    \item \ref{Metrics_1} Mean Cosine Alignment (MCA)
    \item \ref{Metrics_2} Flow Magnitude Variance (FMV)
    \item \ref{Metrics_3} Global Cosine Similarity
    \item \ref{Metrics_4} Global Magnitude Ratio
\end{itemize}

\item \textbf{D. Additional Experimental Results}
\begin{itemize}
    \item \ref{supp:benchmark_categories} Benchmark Scene Categories
    \item \ref{add_exp_1} Additional Quantitative Evaluation
    \item \ref{add_exp_2} Runtime Analysis
    \item \ref{supp:opacity} Opacity Blending Schedule Analysis
    \item \ref{supp:boundary} Boundary Hole Analysis
    \item \ref{detail_2afc} Details of the 2AFC Human Study
    \item \ref{subsec:human_interactivity} Human Study on Interactivity
    \item \ref{add_exp_4} Additional Qualitative Results
\end{itemize}

\end{itemize}

\newpage
\section{Algorithms}
\label{sec: Algorithms}

\subsection{LivingWorld Interactive Pipeline}
\label{sec: Algorithms_1}

For clarity, we present the overall control loop of LivingWorld
in Algorithm~\ref{algorithm}. The algorithm summarizes the
interactive pipeline for constructing and updating environmental
dynamics, including scene expansion, motion alignment, motion
field construction, and bidirectional motion propagation.

\algnewcommand\algorithmicparallel{\textbf{in parallel do}}
\algnewcommand\algorithmicparallelend{\textbf{end parallel}}

\algdef{SE}[PARALLEL]{Parallel}{EndParallel}[1]{\algorithmicparallel\ #1}{\algorithmicparallelend}

\begin{algorithm}[t]
\small
\caption{LivingWorld interactive pipeline for constructing environmental dynamics}
\label{algorithm}
\begin{algorithmic}[1]

\Statex \textbf{Input:} Initial image $I_0$, motion seeds $\{p_i, h_i\}$
\Statex \textbf{Output:} Scene representation $(\mathcal{G}, F_\theta)$

\State $\mathcal{G} \gets \mathrm{InitScene}(I_0)$
\State $S_{\mathrm{prev}} \gets \emptyset$
\State $F_\theta \gets \mathrm{InitMotionField}()$

\Parallel{Thread 1: real-time rendering}
\While{running}
    \State $I_{\mathrm{rend}} \gets \mathrm{Render}(\mathcal{G}, F_\theta, t)$
    \Comment Includes trajectory propagation and opacity
\EndWhile
\EndParallel

\Parallel{Thread 2: environmental dynamics update}
\While{interactive update}

    \State $(\mathcal{G}, I_{\mathrm{new}}, D_{\mathrm{new}}) 
        \gets \mathrm{SceneExpansion}(\mathcal{G})$

    \State $M \gets \mathrm{SAM}(I_{\mathrm{new}}, \{p_i\})$
    \Comment Seed-guided motion masks

    \State $F_{2D} \gets \mathrm{EulerianFlow}(I_{\mathrm{new}}, M, \{h_i\})$
    \Comment Direction-guided motion prediction

    \State $S_i \gets \mathrm{Lift2Dto3D}(F_{2D}, D_{\mathrm{new}})$

    \If{$S_{\mathrm{prev}} \neq \emptyset$}

        \State $(S_i^{m}, S_{\mathrm{prev}}^{m})
            \gets \mathrm{ReprojectionMatch}(S_i, S_{\mathrm{prev}})$

        \State $(R,s) \gets \mathrm{KabschAlign}(S_i^{m}, S_{\mathrm{prev}}^{m})$

        \State $(R,s) \gets \mathrm{RefineAlignment}(R,s,S_i^{m},S_{\mathrm{prev}}^{m})$

        \State $S_{\mathrm{new}} \gets \mathrm{SelectUnmatched}(S_i, S_i^{m})$

        \State $S_{\mathrm{new}} \gets \mathrm{ApplyTransform}(S_{\mathrm{new}}, R, s)$

    \Else

        \State $S_{\mathrm{new}} \gets S_i$

    \EndIf

    \State $S_{\mathrm{prev}} \gets S_{\mathrm{prev}} \cup S_{\mathrm{new}}$

    \State $F_\theta \gets \mathrm{TrainHashMotionField}(F_\theta, S_{\mathrm{prev}})$

\EndWhile
\EndParallel

\end{algorithmic}
\end{algorithm}

\section{Implementation Details} \label{sec:impl_main}
\subsection{Reprojection-based Correspondence Construction}
\label{supp:impl_correspondence}

To align newly predicted scene flow samples with previously accumulated
motion estimations, we establish spatial correspondences using a
reprojection-based matching strategy.

Let $\pi(\cdot)$ denote the camera projection function that maps a
3D point to image coordinates.
Let $a$ and $b$ index individual scene-flow samples in the current
and previously accumulated sets, respectively.
Given the 3D positions associated with the current scene flow
$\mathbf{S}_i$ and the previously accumulated flows
$\mathbf{S}_{\mathrm{prev}}$, we project both point sets into the
current camera view:

\[
(u_{\mathrm{cur}}^{(a)}, v_{\mathrm{cur}}^{(a)})
=
\mathrm{round}\!\left(\pi(\mathbf{x}_{\mathrm{cur}}^{(a)})\right),
\qquad
(u_{\mathrm{prev}}^{(b)}, v_{\mathrm{prev}}^{(b)})
=
\mathrm{round}\!\left(\pi(\mathbf{x}_{\mathrm{prev}}^{(b)})\right).
\]
Here, $\mathbf{x}_{\mathrm{cur}}^{(a)}$ and
$\mathbf{x}_{\mathrm{prev}}^{(b)}$ denote the 3D positions associated
with the $a$-th and $b$-th scene-flow samples in the current and
previously accumulated sets, respectively.
Pixel coordinates are discretized using rounding to match the
implementation.
Correspondences are obtained by identifying samples that share the
same pixel location in the image plane:

\[
\mathcal{M} =
\left\{
(a,b)
\;\middle|\;
(u_{\mathrm{cur}}^{(a)}, v_{\mathrm{cur}}^{(a)})
=
(u_{\mathrm{prev}}^{(b)}, v_{\mathrm{prev}}^{(b)})
\right\}.
\]
The matched scene flow subsets are therefore defined as

\[
S_i^{m} =
\{S_i^{(a)} \mid (a,b) \in \mathcal{M}\},
\qquad
S_{\mathrm{prev}}^{m} =
\{S_{\mathrm{prev}}^{(b)} \mid (a,b) \in \mathcal{M}\}.
\]
These matched flow pairs are subsequently used for estimating the
alignment transformation between the two motion fields.
If reliable correspondences are not available due to limited view overlap, the newly observed scene-flow samples are directly integrated into the accumulated set without performing alignment.

\subsection{Alignment Initialization and Refinement}
\label{supp:impl_alignment}

Given the matched scene-flow subsets $S_i^{m}$ and
$S_{\mathrm{prev}}^{m}$ obtained from the correspondence set
$\mathcal{M}$, we estimate a global similarity transform that aligns
the motion vectors of the current view with the previously accumulated
scene flow.
The alignment consists of a rotation $\mathbf{R} \in SO(3)$ and a
uniform scale $s$.
We formulate the alignment as the following least-squares problem:

\[
\arg\min_{\mathbf{R}, s}
\sum_{(a,b)\in\mathcal{M}}
\left\|
\mathbf{S}_{\mathrm{prev}}^{(b)}
-
s\mathbf{R}\mathbf{S}_i^{(a)}
\right\|^2 .
\]

\noindent
\textbf{Alignment initialization.}
We first compute an initial estimate of the rotation using the
\emph{Kabsch algorithm}~\cite{kabsch1976solution}, which provides the
optimal rotation matrix that minimizes the squared error between the
two matched flow sets in a closed-form manner.
The scale parameter is initialized using a least-squares estimate
based on the magnitudes of the matched motion vectors.
This initialization provides a fast and stable estimate of the
global motion transformation.

\noindent
\textbf{Alignment refinement.}
To further reduce residual discrepancies caused by noisy flow
predictions and depth estimation errors, we perform a lightweight
gradient-based refinement step.
The rotation is parameterized using an axis--angle representation,
which allows the rotation matrix $\mathbf{R}$ to be updated through
standard gradient optimization.

Starting from the initialized $(\mathbf{R}, s)$, we minimize the same
alignment objective using stochastic gradient descent (SGD).
In our implementation, the optimization is performed for
300 iterations with a learning rate of $1\times10^{-1}$.
Since the refinement operates only on the sparse matched flow
samples, the computational overhead is negligible while improving
the stability and accuracy of the alignment.

\subsection{Hash-based Motion Field}
\label{supp:impl_hash}

To represent spatially continuous environmental dynamics,
we learn a neural motion field $F_\theta : \mathbb{R}^3 \rightarrow
\mathbb{R}^3$ that maps a 3D position to a motion vector.
The motion field is trained using the union of the previously
accumulated scene-flow samples $S_{\mathrm{prev}}$ and the
newly estimated flows at the current step.
This training strategy ensures that the learned motion field
remains consistent with both previously observed motion and
newly incorporated scene-flow samples, preventing drift during
incremental scene expansion.

\noindent
To efficiently model spatial variations in the motion field,
we adopt a multi-resolution hash-grid encoding similar to
Instant-NGP~\cite{muller2022instant}.
Each input position $\mathbf{x}$ is first normalized according to
the spatial extent of the observed scene.
The normalization parameters are computed from the bounding box
of the accumulated 3D points associated with the scene-flow samples
$S_{\mathrm{prev}}$.
Specifically, we estimate the center $\mathbf{c}$ and half-extent
$\mathbf{b}$ of the point set and normalize positions as
\[
\tilde{\mathbf{x}} =
\frac{\mathbf{x}-\mathbf{c}}{\mathbf{b}} .
\]
The normalized coordinates are then mapped to $[0,1]^3$ before being
passed to the hash-grid encoder.
The encoder consists of 16 resolution levels with
4 features per level and a hash map size of $2^{19}$.
The base grid resolution is set to 16 and grows geometrically
with a scale factor of 1.5 across levels.

\noindent
\textbf{Motion prediction network.}
The encoded features are processed by a lightweight
multi-layer perceptron that predicts the 3D motion vector
at each spatial location:
\[
F_\theta(\mathbf{x}) =
\text{MLP}(\text{Enc}(\tilde{\mathbf{x}})).
\]
The network outputs a 3-dimensional motion vector
representing the local velocity at position $\mathbf{x}$.

\noindent
\textbf{Training objective.}
The motion field is trained to regress the accumulated
scene-flow samples by minimizing a mean-squared error loss:

\[
\mathcal{L}_{\text{motion}} =
\sum_i
\left\|
F_\theta(\mathbf{x}_i) -
\mathbf{s}_i
\right\|^2,
\]
where $\mathbf{x}_i$ and $\mathbf{s}_i$ denote the position and corresponding scene-flow vector of the $i$-th sample
in $S_{\mathrm{prev}}$.

\noindent
\textbf{Optimization details.}
The network is trained using the Adam optimizer with a learning rate of $10^{-2}$ for 100 iterations.
In practice, the lightweight network converges quickly, and no extensive optimization is required.
Because the number of scene-flow samples is moderate
and the network is lightweight, the training process
typically converges within a few seconds and does not affect the interactive performance of the system.

\section{Motion Consistency Metrics}
\label{Metrics}

To quantitatively evaluate motion consistency in the ablation study
of the Geometry-Aware Alignment Module, we use both local and global
consistency metrics. Local consistency is measured by Mean Cosine
Alignment (MCA) and Flow Magnitude Variance (FMV), which evaluate
neighboring motion vectors. Global consistency is measured by cosine
similarity and magnitude ratio between spatially distant point pairs
within the same motion region, including newly expanded areas.

\subsection{Mean Cosine Alignment (MCA)}
\label{Metrics_1}

Mean Cosine Alignment measures the local directional consistency of
neighboring motion vectors. For each scene-flow sample $\mathbf{s}_p$,
we compute the $K$-nearest neighbors $\mathcal{N}(p)$ based on Euclidean
distance in 3D space. The flow vectors are first normalized as
\[
\hat{\mathbf{s}}_p =
\frac{\mathbf{s}_p}{\|\mathbf{s}_p\|}.
\]
MCA is defined as
\[
\text{MCA} =
\frac{1}{N}
\sum_{p=1}^{N}
\left(
\frac{1}{K}
\sum_{j \in \mathcal{N}(p)}
\hat{\mathbf{s}}_p \cdot \hat{\mathbf{s}}_j
\right).
\]
A higher MCA indicates stronger local directional coherence.

\subsection{Flow Magnitude Variance (FMV)}
\label{Metrics_2}

Flow Magnitude Variance measures the local smoothness of motion
magnitudes. Let $m_p = \|\mathbf{s}_p\|$ denote the magnitude of the
scene flow at point $p$. Using the same neighborhood structure
$\mathcal{N}(p)$, FMV is defined as
\[
\text{FMV} =
\frac{1}{N}
\sum_{p=1}^{N}
\left(
\frac{1}{K}
\sum_{j \in \mathcal{N}(p)}
(m_p - m_j)^2
\right).
\]
A lower FMV indicates smoother local variations in motion magnitude.

\subsection{Global Cosine Similarity}
\label{Metrics_3}

To measure global directional consistency, we sample spatially distant
point pairs $(p,q)$ within the same motion region, including newly
expanded areas. For each pair, we compute the cosine similarity between
their normalized scene-flow vectors:
\[
\text{Global Cosine} =
\frac{1}{|\mathcal{P}|}
\sum_{(p,q)\in\mathcal{P}}
\frac{\mathbf{s}_p \cdot \mathbf{s}_q}
{\|\mathbf{s}_p\|\|\mathbf{s}_q\|},
\]
where $\mathcal{P}$ denotes the set of sampled distant point pairs.
A higher value indicates that motion directions remain consistent
across distant regions of the generated world.

\subsection{Global Magnitude Ratio}
\label{Metrics_4}

To measure global magnitude consistency, we use the magnitude ratio
between the same distant point pairs:
\[
\text{Mag. Ratio} =
\frac{1}{|\mathcal{P}|}
\sum_{(p,q)\in\mathcal{P}}
\frac{\min(\|\mathbf{s}_p\|,\|\mathbf{s}_q\|)}
{\max(\|\mathbf{s}_p\|,\|\mathbf{s}_q\|)}.
\]
This value lies in $[0,1]$, where a value closer to 1 indicates more
consistent motion strength across spatially distant regions.

\newpage
\section{Additional Experiment Results}
\label{add_exp}

\subsection{Benchmark Scene Categories}
\label{supp:benchmark_categories}

The benchmark used in the main paper consists of 60 scenes covering four categories of environmental dynamics: clouds, water, smoke/fog, and fire, with 15 scenes per category.
Fig.~\ref{fig:scene_categories} shows representative examples from each category.

\begin{figure}[t]
\centering
\includegraphics[width=\textwidth]{figure/scene_categories.pdf}
\caption{
Benchmark scene categories used for the main quantitative evaluation.
}
\label{fig:scene_categories}
\end{figure}

\subsection{Additional Quantitative Evaluation on In-the-Wild Dataset}
\label{add_exp_1}

To further evaluate the generalization of the proposed method beyond the curated scenes used in the main paper, we conduct additional quantitative experiments on scenes collected from Pexels~\cite{pexels} and Unsplash~\cite{unsplash}.
These scenes contain diverse natural environments and provide
a complementary evaluation setting for dynamic scene generation.
We follow the same evaluation protocol used in the main paper
and compute the GPT-4o based metrics, PhysReal and PhotoReal,
which measure physical plausibility and visual realism of the
generated dynamic scenes.
Table~\ref{tab:in-the-wild} reports the results on these additional
in-the-wild scenes. The proposed framework achieves consistently
strong performance across both metrics, demonstrating robust
motion generation across diverse natural scenes.

\begin{table}[t]
\centering
\footnotesize
\setlength{\tabcolsep}{5pt}
\renewcommand{\arraystretch}{1.05}
\resizebox{0.8\columnwidth}{!}{
\begin{tabular}{@{}l|l|c|c|c@{}}
\toprule
\textbf{Category} & \textbf{Method} & \textbf{PhysReal (↑)} & \textbf{PhotoReal (↑)} & \textbf{Time (s)}\\
\midrule
\multirow{3}{*}{Video Gen.} 
& Veo 3.1~\cite{wiedemer2025video} & 0.80 & 0.82 & 140  \\
& CogVideoX~\cite{yang2024cogvideox} & 0.75 & 0.79 & 1404  \\
& Tora~\cite{zhang2025tora} & 0.79 & 0.82 & 546 \\
\midrule
\multirow{2}{*}{4D Scene}
& 4DGS-Cinemagraphy~\cite{jin2025optimizing} & 0.81 & 0.80 & 2100 \\
& \textbf{LivingWorld (Ours)} & \textbf{0.82} & \textbf{0.83} & \textbf{12} \\
\bottomrule
\end{tabular}
}
\caption{Quantitative evaluation on additional in-the-wild dataset collected from publicly available images. 
PhysReal and PhotoReal are GPT-4o based metrics that measure the physical plausibility and visual realism of the generated dynamic scenes.}
\label{tab:in-the-wild}
\vspace{-2mm}
\end{table}

\subsection{Runtime Analysis}
\label{add_exp_2}

We report the runtime of the main components of the proposed
pipeline to provide a clearer understanding of the computational
cost of interactive 4D scene generation.
All experiments are conducted on a single NVIDIA RTX 5090 GPU.
During interactive exploration, the system maintains two parallel
processes: (1) real-time rendering of the current dynamic scene,
and (2) incremental scene expansion and motion update.
Table~\ref{tab:runtime} summarizes the runtime of the major
modules in the update pipeline.
The motion update stage in the main paper corresponds to the combined cost of motion estimation, geometry-aware alignment, and hash-field update, which together take approximately 3 seconds.

\begin{table}[h]
\centering
\small
\setlength{\tabcolsep}{6pt}
\begin{tabular}{l c}
\toprule
Module & Runtime (s) \\
\midrule
Scene expansion (outpainting + depth) & 9.0 \\
Motion estimation (Eulerian flow) & 0.2 \\
Motion alignment (Kabsch + refinement) & 0.3 \\
Motion field update (hash-grid training) & 2.5 \\
\midrule
Total update time & $\approx$ 12.0 \\
\bottomrule
\end{tabular}
\caption{Runtime breakdown of the main modules in the proposed pipeline.}
\label{tab:runtime}
\end{table}

\subsection{Opacity Blending Schedule Analysis}
\label{supp:opacity}

\begin{figure}[t]
\centering
\includegraphics[width=0.85\linewidth]{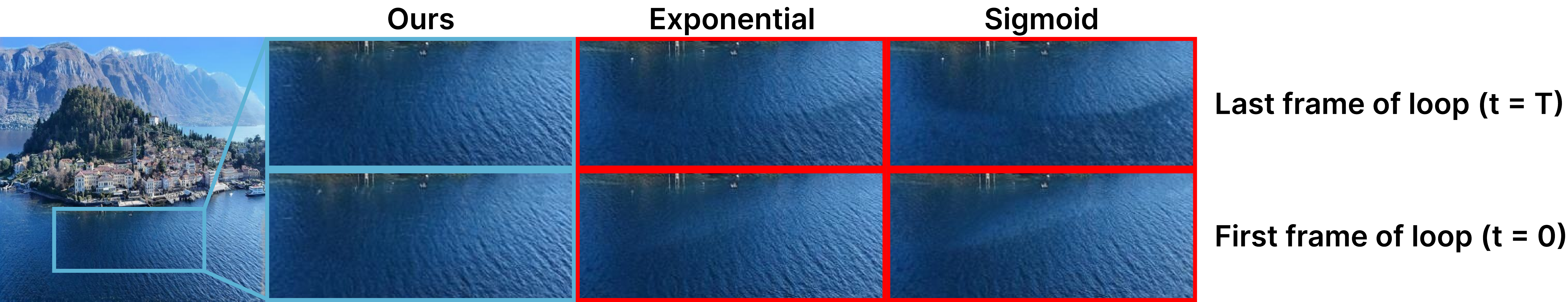}
\caption{
Comparison of opacity blending schedules for looped motion rendering.
Nonlinear schedules can introduce abrupt changes near the loop boundary,
whereas the linear schedule provides a smoother transition between the
forward and backward trajectories.
}
\label{fig:opacity_schedule}
\end{figure}

To construct a temporally seamless looping animation, we linearly blend
the forward and backward trajectories using the opacity schedule
$w(t)=t/T$.
We compare this linear schedule with nonlinear alternatives, including exponential and sigmoid schedules.
As shown in Fig.~\ref{fig:opacity_schedule}, nonlinear schedules produce
abrupt opacity changes near the loop boundary between the last and first
frames, leading to visible temporal discontinuities.
In contrast, the linear schedule provides a smoother transition and
maintains stable opacity throughout the loop.

\subsection{Boundary Hole Analysis}
\label{supp:boundary}

Forward-only motion propagation gradually reduces Gaussian density around
motion boundaries, producing holes as motion accumulates.
To quantify this effect, we measure the percentage of pixels inside the
motion mask that become uncovered after motion propagation.
As shown in Table~\ref{tab:boundary_hole}, bidirectional propagation
reduces the hole rate from 9.8\% to 0.9\%, corresponding to an
approximately 91\% reduction.
Although small gaps may remain in challenging cases, the proposed strategy
substantially improves boundary completeness while maintaining stable
temporal motion.

\begin{table}[t]
\centering
\footnotesize
\setlength{\tabcolsep}{8pt}
\renewcommand{\arraystretch}{1.05}
\begin{tabular}{l|c}
\toprule
\textbf{Method} & \textbf{Hole Rate (\%) $\downarrow$} \\
\midrule
Forward-only propagation & 9.8 \\
Bidirectional propagation (Ours) & \textbf{0.9} \\
\bottomrule
\end{tabular}
\caption{Boundary hole analysis. Hole rate denotes the percentage of uncovered pixels inside the motion mask after motion propagation.}
\label{tab:boundary_hole}
\end{table}

\subsection{Details of the 2AFC Human Study}
\label{detail_2afc}
\begin{figure}[t]
\centering
\includegraphics[width=\linewidth]{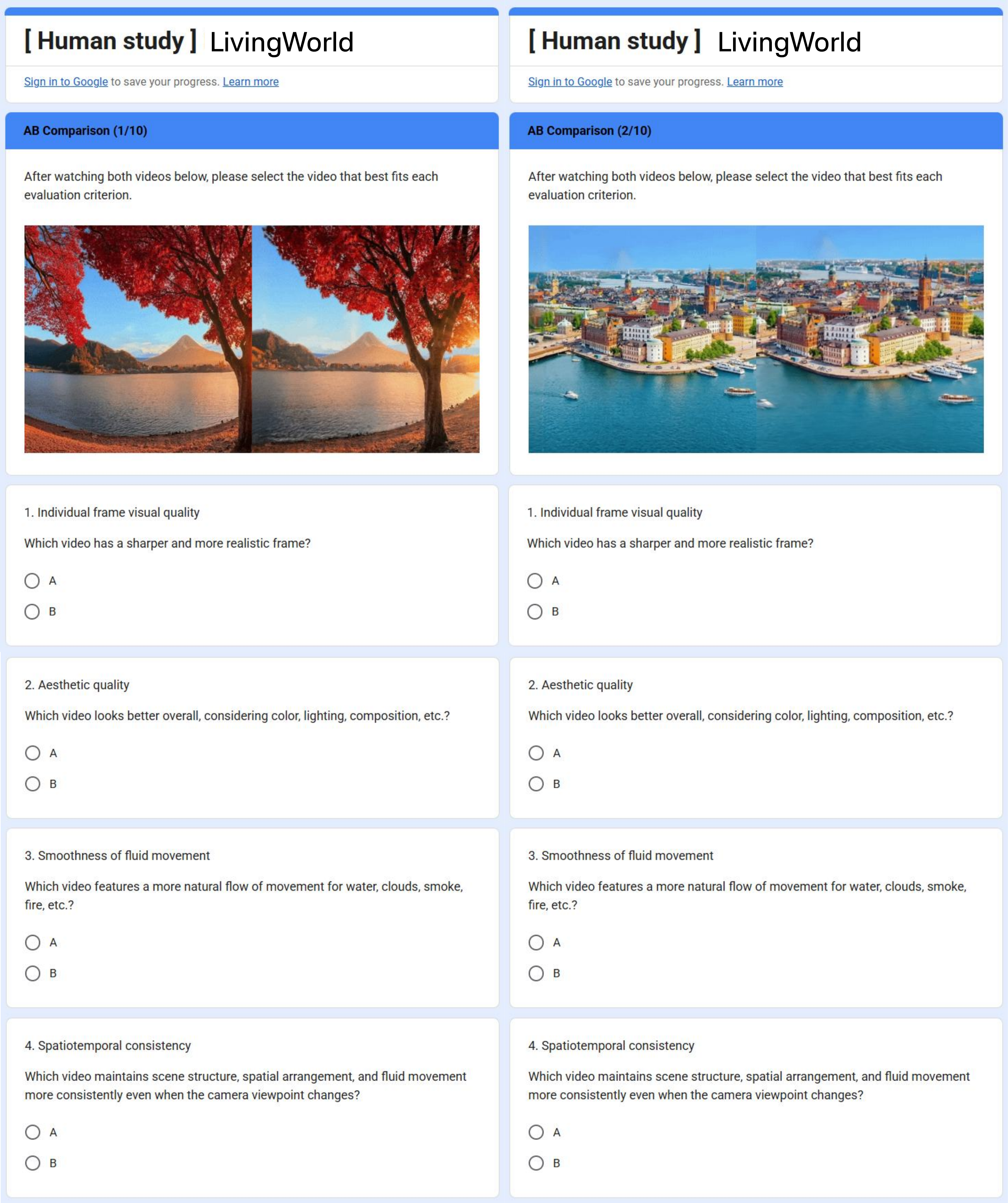}
\caption{
Questionnaire example used in the 2AFC human preference study.
Participants compare two videos generated from the same input and select the preferred result for each evaluation criterion.
}
\label{fig:human_protocol}
\end{figure}

To complement the automatic metrics reported in the main paper, we conduct a 2AFC human preference study using criteria aligned with the VBench evaluation protocol.
A total of 95 participants took part in the study.
For each comparison, participants were presented with two videos generated from the same input image and camera trajectory: one produced by LivingWorld and the other by a baseline method.
The presentation order was randomized to avoid bias.
Participants evaluated the videos according to four criteria:
(1) Imaging Quality,
(2) Aesthetic Quality,
(3) Motion Smoothness, and
(4) Temporal Consistency (Flicker).
Fig.~\ref{fig:human_protocol} shows an example of the questionnaire used in the study.
Each participant evaluated multiple scenes sampled from the benchmark used in the main paper.
The results are reported in the main paper.
LivingWorld is consistently preferred across all baselines, particularly for temporal criteria such as Motion and Flicker.
These preferences are consistent with the automatic evaluation results and provide additional human validation of the temporal coherence of the generated environmental dynamics.

\clearpage

\clearpage

\subsection{Human Study on Interactivity}
\label{subsec:human_interactivity}
\begin{figure}[t]
    \centering
    \includegraphics[width=\textwidth]{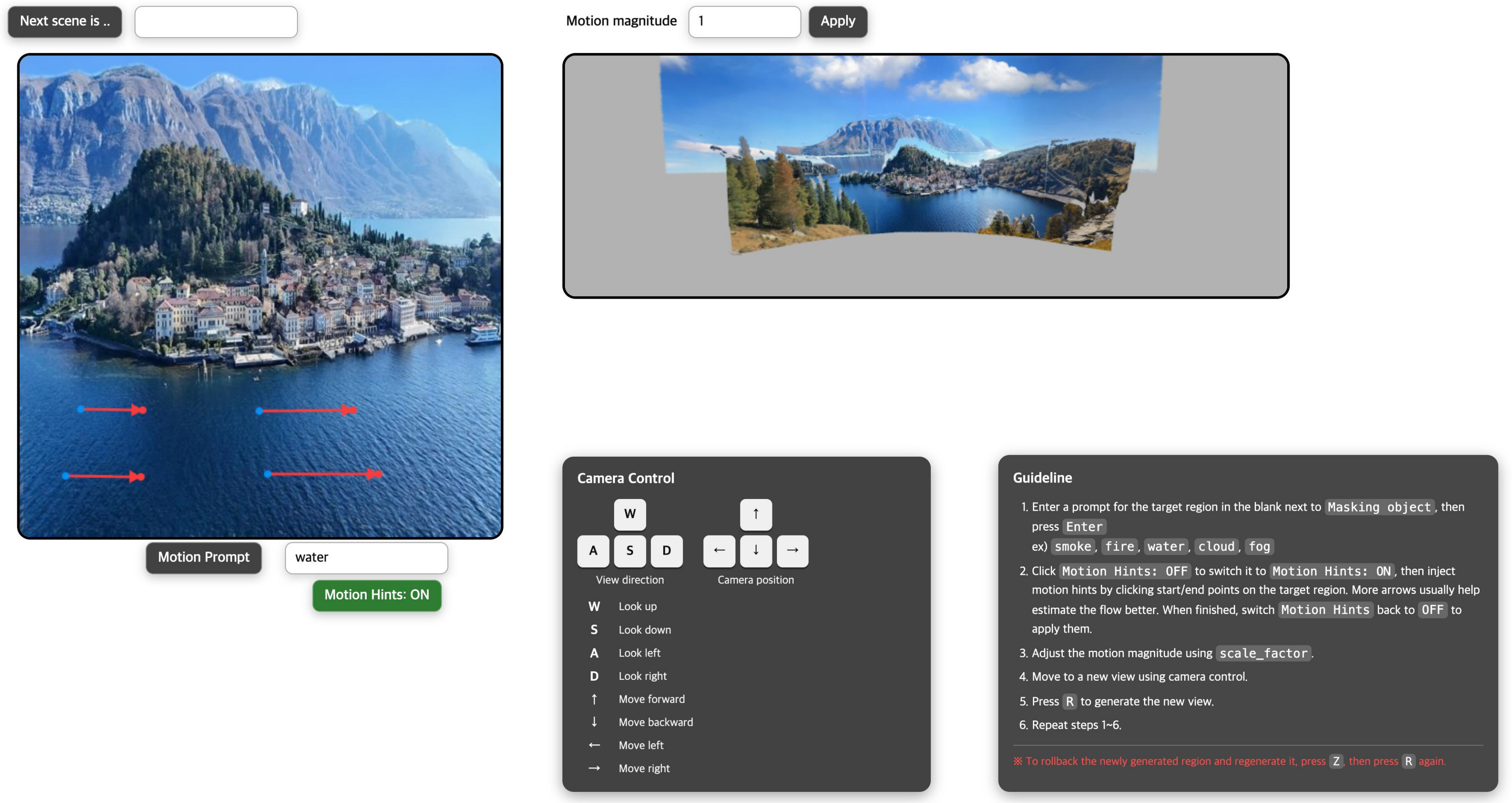}
    \caption{LivingWorld user interface used in the interactivity human study.}
    \label{fig:livingworld_interface}
\end{figure}

Runtime alone does not fully capture the interactivity and usability of the system. We therefore conduct a human study to evaluate how easily participants can use the LivingWorld interface for interactive 4D world generation. A total of 20 participants with design or software-development backgrounds took part in the study and completed the assigned interaction tasks. While the 2AFC human study focuses on the perceptual quality of the generated videos, this study evaluates whether participants can intuitively perform camera control, dynamic region selection, and motion guidance. LivingWorld is built upon a WonderWorld-style interactive world generation interface that supports camera-based scene expansion. Compared with this baseline interface, LivingWorld additionally provides dynamic region selection and motion hint controls. We compare LivingWorld with the baseline interface to assess whether these additional 4D motion controls can be introduced without substantially reducing the convenience of the original interactive world generation workflow. Fig.~\ref{fig:livingworld_interface} shows the LivingWorld user interface used in the study. The interface displays the input image, the current generated world view, camera controls, and motion-guidance tools. Participants can specify dynamic regions through mask prompts, provide motion hints by clicking on the input image, adjust motion magnitude, and move the camera to generate new views.
After completing the assigned tasks, participants evaluated the system using a 7-point Likert scale in terms of usability, controllability, and usefulness. Fig.~\ref{fig:google_form_interactivity} shows the questionnaire used in this study, and Table~\ref{tab:interactivity_study} summarizes the resulting Likert-scale ratings. The results show that LivingWorld achieves positive usability ratings comparable to the WonderWorld-style baseline interface while providing additional controls for dynamic 4D world generation. Although the added motion controls slightly increase interaction complexity, participants still rated the system positively across all criteria. This suggests that LivingWorld can support interactive camera navigation and motion guidance without substantially compromising usability.

\begin{table}[t]
\centering
\begin{tabular}{l|c|c}
\toprule
Metric & LivingWorld & Baseline \\
\midrule
Usability & 5.65($\pm$1.25) & 5.80($\pm$0.71) \\
Controllability & 5.90($\pm$1.42) & 6.20($\pm$1.04) \\
Usefulness & 6.00($\pm$1.21) & 6.20($\pm$1.04) \\
Average & 5.85($\pm$1.30) & 5.93($\pm$0.68) \\
\bottomrule
\end{tabular}
\caption{Interactivity human study using a 7-point Likert scale.}
\label{tab:interactivity_study}
\end{table}

\begin{figure}[t]
    \centering
    \includegraphics[width=\textwidth]{figure/google_form_interactivity.pdf}
    \caption{
    Questionnaire used in the interactivity human study.
    Participants evaluate the system in terms of usability, controllability, and usefulness using a 7-point Likert scale.
    }
    \label{fig:google_form_interactivity}
\end{figure}

\clearpage

\subsection{Additional Qualitative Results}
\label{add_exp_4}

\begin{figure*}[!t]
\centering
\includegraphics[width=\textwidth]{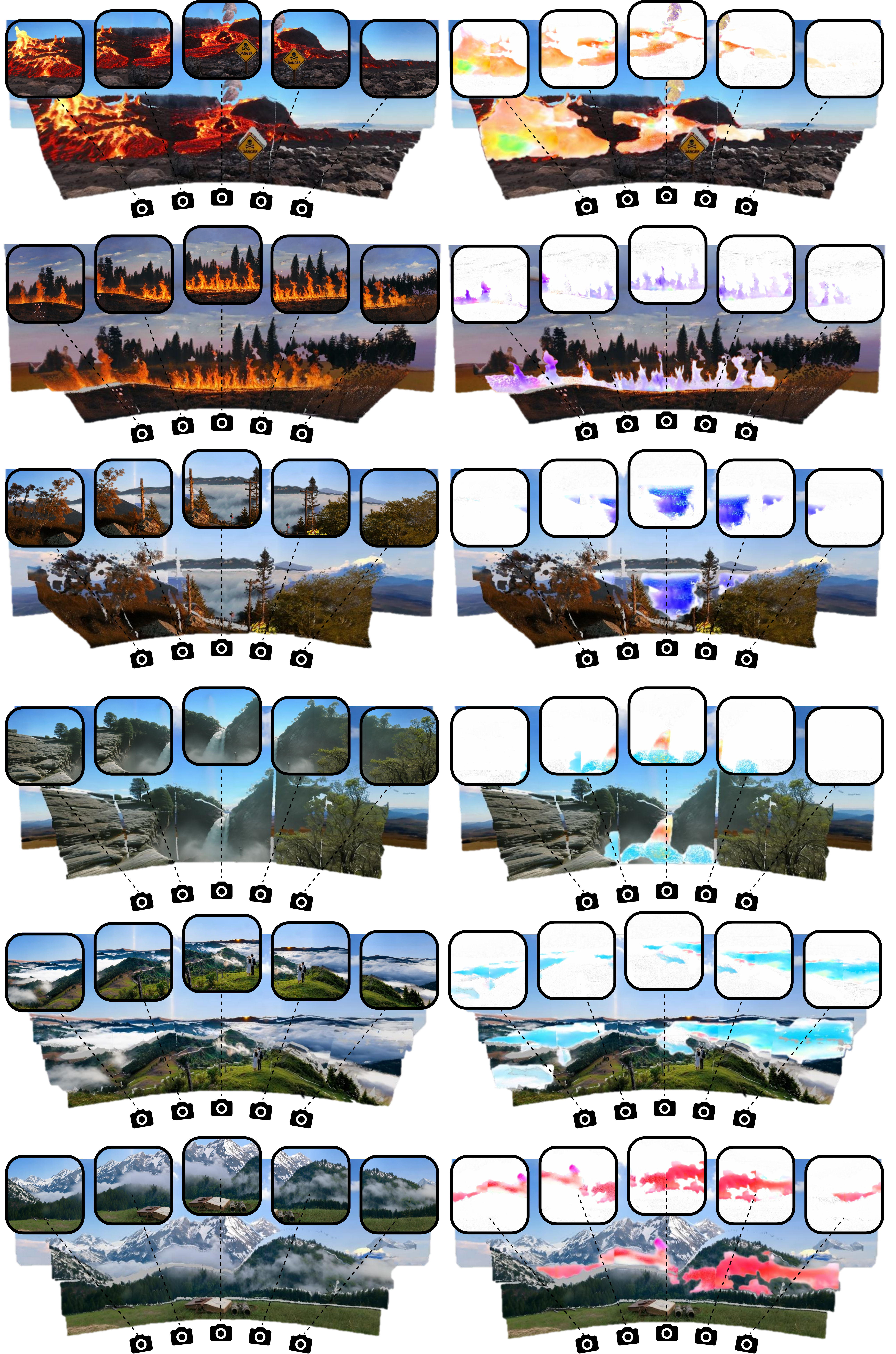}
\caption{Additional qualitative results on curated scenes.}
\label{fig:qualitative_b}
\end{figure*}

\begin{figure*}[!t]
\centering
\includegraphics[width=\textwidth]{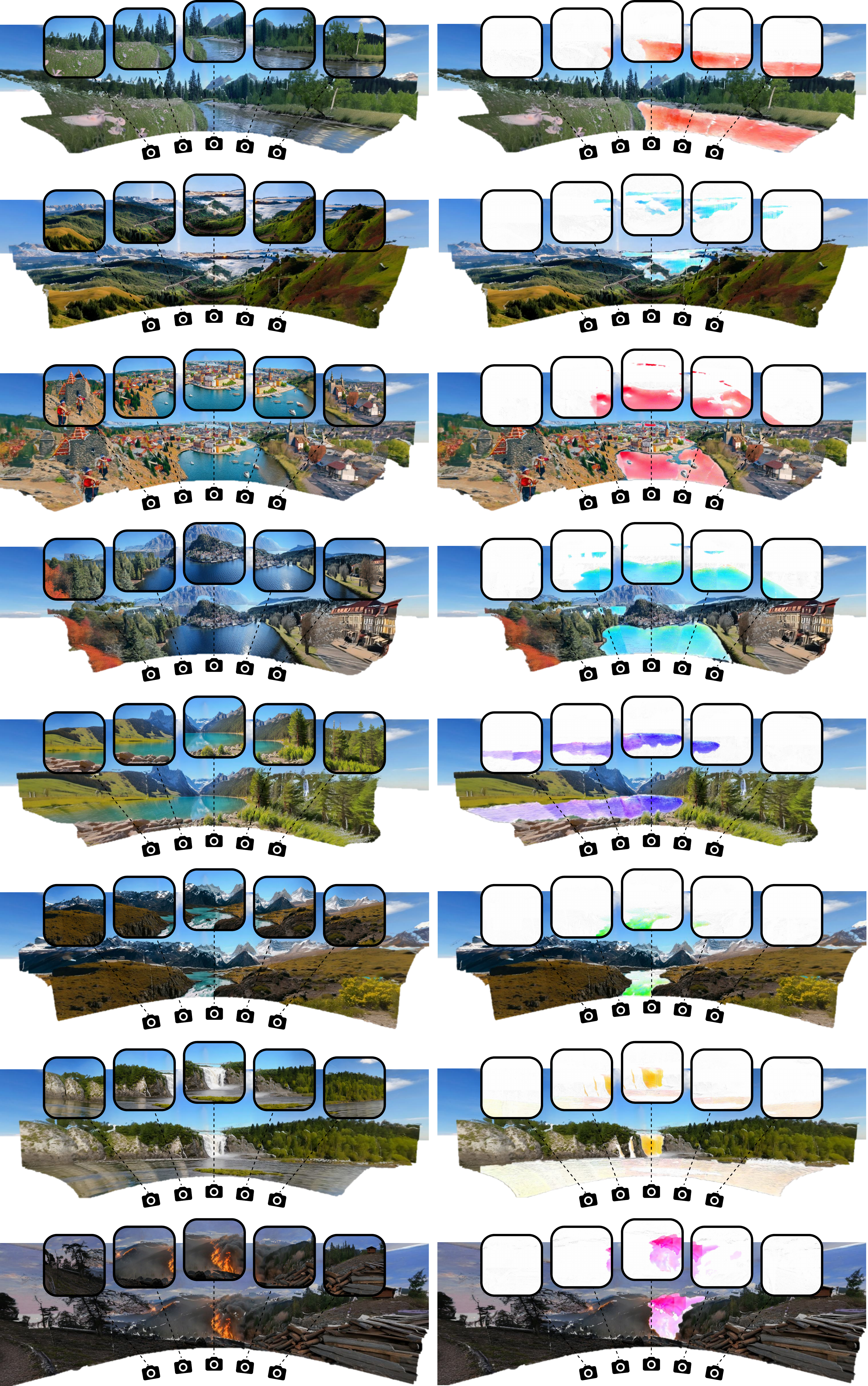}
\caption{Additional qualitative results on in-the-wild scenes.}
\label{fig:qualitative_c}
\end{figure*}

\begin{figure*}[!t]
\centering
\includegraphics[width=\textwidth]{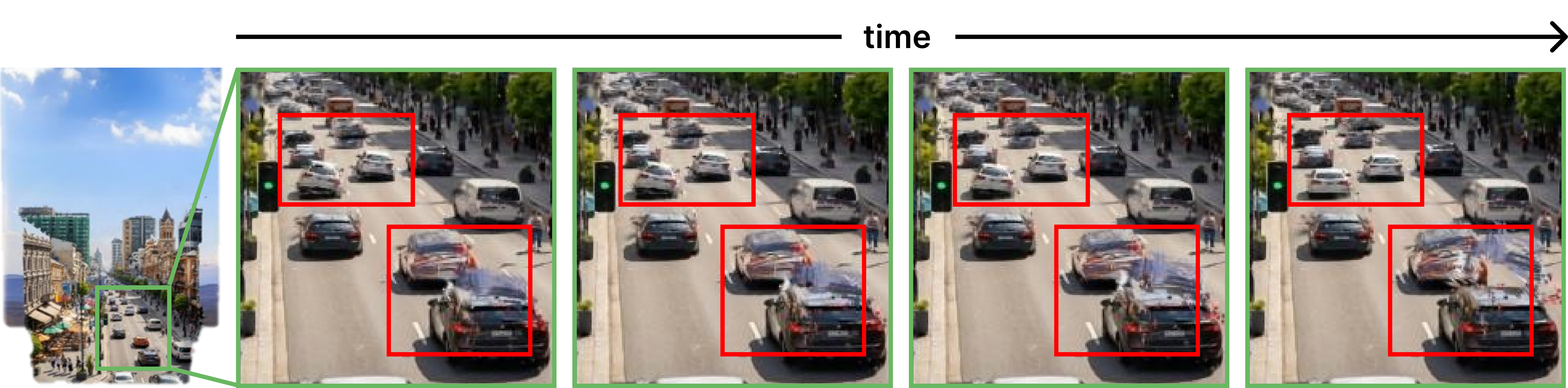}
\caption{\textbf{Rigid-body motion is outside the primary scope of our framework.}
Applying motion to rigid objects may introduce geometric distortions, since the proposed method models dynamics as continuous spatial velocity fields designed primarily for environmental dynamics.}
\label{fig:failure_car}
\end{figure*}

Figs.~\ref{fig:qualitative_b} and~\ref{fig:qualitative_c} present additional qualitative examples,
including scenes sampled from both the curated dataset used in the main paper
and additional in-the-wild dataset.
These examples exhibit diverse environmental dynamics such as clouds,
smoke, and fire.
The results demonstrate that the proposed framework can generate
spatially coherent motion across a wide range of environments beyond
those shown in the main paper.
Fig.~\ref{fig:failure_car} illustrates a case where motion is applied to a rigid object.
Since our framework models dynamics as a continuous spatial velocity field primarily designed for environmental motion, minor geometric distortions may appear in strictly rigid-body motion.

\end{document}